\documentclass{article}
\usepackage{amsmath}
\usepackage{psfrag}
\usepackage{hyperref}
\usepackage{import}
\usepackage{subfigure}
\usepackage{ragged2e}

\usepackage{algorithm}
\usepackage[noend]{algpseudocode}

\usepackage{xcolor}			
\usepackage{color}			
\usepackage{transparent}

\usepackage[normalem]{ulem}

\usepackage{setspace}

\newcommand{\bs}{\boldsymbol}
\def\RR{ \mathbb R}
\newcommand{\refeq}[1]{Equation \eqref{#1}}

\newcommand{\ee}{\end{equation}}
\newcommand{\be}{\begin{equation}}
\newcommand{\ec}{\end{center}}
\newcommand{\bc}{\begin{center}}
\newcommand{\eea}{\end{eqnarray}}
\newcommand{\bea}{\begin{eqnarray}}
\newcommand{\bd}{\begin{description}}
\newcommand{\ed}{\end{description}}
\newcommand{\bi}{\begin{itemize}}
\newcommand{\ei}{\end{itemize}}
\newcommand{\pa}{\partial}
\newcommand\bphi{\boldsymbol{\phi}}
\newcommand\bpsi{\boldsymbol{\psi}}

\newcommand{\bx}{\bs{x}}

\newcommand{\by}{\bs{y}}

\newcommand{\bxx}{\bs{X}}

\newcommand{\byy}{\bs{Y}}

\newcommand{\bt}{\bs{\theta}}

\usepackage{amssymb}
\usepackage{PRIMEarxiv}

\usepackage[utf8]{inputenc} 
\usepackage[T1]{fontenc}    
\usepackage{hyperref}       
\usepackage{url}            
\usepackage{booktabs}       
\usepackage{amsfonts}       
\usepackage{nicefrac}       
\usepackage{microtype}      
\usepackage{lipsum}
\usepackage{graphicx}
\graphicspath{{media/}}     

\title{Physics-Aware Neural Implicit Solvers for multiscale, parametric PDEs with applications in heterogeneous media.
\thanks{\textit{Corresponding author\\
Email addresses: matthaios.chatzopoulos@tum.de (Matthaios Chatzopoulos), \\
p.s.koutsourelakis@tum.de (Phaedon-Stelios Koutsourelakis)
}} 
}

\author{
  Matthaios Chatzopoulos$^a$, Phaedon-Stelios Koutsourelakis$^{a, b, *}$ \\
  $^a$ Technical University of Munich, Professorship of Data-driven Materials Modeling,\\ School of
Engineering and Design, Boltzmannstr. 15, 85748 Garching, Germany \\
$^b$ Munich Data Science Institute (MDSI - Core member), Garching, Germany}

\begin{document}
\maketitle

\begin{abstract}
We propose Physics-Aware Neural Implicit Solvers (PANIS), a novel, data-driven framework for learning surrogates for parametrized Partial Differential Equations (PDEs). 
It consists of a  probabilistic, learning objective in which  weighted residuals are used to probe the PDE and provide a source of {\em virtual} data i.e. the actual PDE never needs to be solved. This is combined with a physics-aware implicit solver that consists of a much coarser, discretized version of the original PDE, which provides the requisite information bottleneck for high-dimensional problems and enables generalization in out-of-distribution settings (e.g. different boundary conditions).
We demonstrate its capability in the context of random heterogeneous materials where the input parameters represent the material microstructure. We extend the framework to multiscale problems and show that a surrogate can be learned for the effective (homogenized) solution without ever solving the reference problem. We further demonstrate how the proposed framework can accommodate and generalize several existing learning objectives and architectures while yielding probabilistic surrogates that can quantify predictive uncertainty.
\end{abstract}


\section{Introduction}
\label{introduction}

Parametric PDEs appear in a wide variety of problems of engineering relevance, and their repeated solution under different parametric values in the context of many-query applications represents a major computational roadblock in achieving analysis and design objectives.
Perhaps one of the most challenging applications, which lies at the core of this investigation, is encountered  in the context of (random) heterogeneous media in which microstructural details determine their macroscopic properties \cite{torquato2002random}. 
These are found in  a multitude of engineering applications, such as aligned and chopped fiber composites, porous membranes, particulate composites, cellular solids, colloids, microemulsions, concrete
\cite{torquato2002random}. 
Their microstructural  properties can vary, most often randomly,    at multiple length-scales \cite{stefanou_random_2021}. Capturing this variability requires, in general, very high-dimensional representations and very fine discretizations, which in turn imply a significant cost for each solution of the governing PDEs in order to predict their response \cite{yabansu_digital_2020}.
Being able to efficiently obtain accurate solutions under varying microstructures represents a core challenge that can enable the solution of various forward analysis problems such as uncertainty quantification \cite{panchal2013key,arroyave_systems_2019}. More importantly, however, it is of relevance in the context of inverse design where one attempts to identify (families of) microstructures that achieve extremal or target properties \cite{lee_fast_2021}. 
While several different tools come into play, {\em data-driven} strategies, to which our contribution belongs, have risen into prominence in recent years \cite{agrawal_perspective_2016,kalidindi2015hierarchical} as in many cases  they have produced high-throughput, forward-model surrogates which  are essential for inverting the microstructure-to-property link \cite{curtarolo_high-throughput_2013}.

One way to categorize pertinent  methods is based on the learning objectives employed which range between purely data-based to physics-informed. In the first category belong methods that rely on {\em labeled} data, i.e.   input-output pairs \cite{ yang_establishing_2019, lu2019deeponet, li2020neural,li2020fourier,you_learning_2022} and cast the problem as a supervised learning task. Associated models and tools have reached a high-level of maturity and sophistication in the machine learning community and exhibit a comparatively  faster and more stable training \cite{karniadakis2021physics}.  If the models employed, which usually take the form of a Deep Neural Network (DNN), are agnostic of the physical problem, their predictive accuracy is generally dependent on the amount of training data. However, unlike  typical machine learning applications, as e.g. in language or vision,   where data is abundant and inexpensive, in parametric PDEs, data acquisition is de facto expensive, and the reduction of the requisite number of PDE-solves one of the primary objectives \cite{koutsourelakis_special_2016}.  

On the other side of the spectrum lie physics-informed learning protocols \cite{raissi2017physics,wang2021learning,vadeboncoeur2023random}. These employ the PDE itself, usually in the form of collocation-type residuals, in the training loss, although energy functionals have also  been  utilized  \cite{zhu2019physics,rixner2021probabilistic}. As such, they do not require any PDE-solves and exhibit better generalization performance, but empirical results have shown that the use of labeled data can improve their performance \cite{zhu2019physics}. 
In the works of \cite{raissi2017physics}, \cite{yang2019adversarial}, \cite{kaltenbach2020incorporating}, \cite{rixner2021probabilistic}, \cite{raissi2018hidden}, \cite{yu2018deep} the loss function used for training the model  consisted of both labeled data and residuals of the governing PDE. A detailed taxonomy \cite{kim2021knowledge} and review regarding the application of deep, physics-informed models for solving  pertinent problems can be found in 
\cite{karniadakis2021physics}.
The relative weights of the  residual-based, data-driven and regularization terms in the loss are generally ad hoc and fine-tuned through trial and error. Furthermore, no quantification of the informational content of the residuals employed is directly available, and in general, a large number of collocation points is needed.  

Another aspect of the problem that plays a pivotal role is the dimension of the parametric input. 
In cases where this is low to medium, generic,   traditional architectures such as 
Gaussian Processes \cite{BILIONIS20125718}, 
exhibit good results. 
Dimensionality reduction offers an avenue for overcoming difficulties in high-dimensional problems and several such strategies have been developed \cite{quarteroni2015reduced, hesthaven2016certified,haasdonk2017reduced,generale_reduced-order_2021}. 
Since, in general, dimensionality reduction is performed in a separate step,  if the lower-dimensional features identified are not affecting the response/properties, their utility is limited \cite{hashemi_feature_2021}. Furthermore, most of the attention has been directed towards reducing the dimensionality of the entities involved \cite{kalidindi_feature_2020} rather than reducing, or better yet, coarse-graining, the physical models \cite{wen_multiscale_2012}.

Recent advances in deep learning \cite{Goodfellow-et-al-2016} have triggered a big effort towards the utilization of Deep Neural Networks for solving efficiently such problems  \cite{han2018solving,sirignano2018dgm,li2017deep}. Combining DNNs with dimensionality reduction techniques has also  been employed  in problems involving heterogeneous media as in  
\cite{yang2019conditional, zhu2018bayesian,mo2019deep}. 
When the input is infinite-dimensional (i.e. a function), one of  the latest and most promising advancements are the so-called Neural Operators \cite{kovachki2021neural}, which attempt to approximate directly the map to the PDE-solution, i.e.  a function to function mapping between  Banach spaces. Typical examples include the  Deep Operator Network (DeepONet) \cite{lu2019deeponet} (and its physics-informed version in  \cite{wang2021learning}),  the  Graph Neural Operator (GNO) \cite{li2020neural}, the Fourier Neural Operator (FNO)  
\cite{li2020fourier,li2021physics}, the Wavelet Neural Operator 
 
\cite{tripura2022wavelet} and the Convolutional Neural Operator \cite{raonic2024convolutional}. 
While such methods promise resolution-independent results, they sample the input function on a given grid and employ a DNN to parametrize the PDE solution. As shown in \cite{fanaskov_spectral_2023} this can lead to systematic bias and aliasing errors which could be overcome by using truncated (i.e. finite-dimensional) Chebyshev or Fourier series for both domain and co-domain.

We note finally that most deep-learning-centered strategies have been based on transferring or improving NN architectures that parametrize the   input-output map. Fewer efforts have attempted to incorporate  physical inductive biases \cite{cranmer_discovering_2020} in those architectures by e.g. endowing them with invariance/equivariance properties when those are available \cite{kohler2020equivariant}.

Despite notable progress, very few models are capable of producing probabilistic predictive estimates \cite{grigo_physics-aware_2019,rixner2021probabilistic,garg2022variational,vadeboncoeur2023fully}. 
This can be achieved by learning/approximating the posterior (given labeled data) of the model parameters, which is a difficult task given the millions or even billions of parameters that DNNs possess.   Much fewer efforts attempt to learn directly a predictive posterior of the solution \cite{rixner2021probabilistic}. We believe, nevertheless, that this is very important for any data-driven scheme (even more so for over-parametrized models) given that a finite number of data points can be employed and, in fact, reducing the requisite information extracted in the form of input-output pairs or otherwise from the PDE, is one of the main objectives. Information loss that can give rise to predictive uncertainty also arises due to the processes of dimensionality reduction or model compression/reduction, which should be quantified and propagated in the estimates.

In this work, we revisit the way information from the governing PDE is ingested 
 as well as the architecture employed to capture the sought input-output map \cite{zhu2019physics}.
 In section \ref{sec:Methodology}, we propose Physics-Aware Neural Implicit Solvers (PANIS), a novel, data-driven framework  which consists of a  probabilistic, learning objective in which  weighted residuals (section \ref{sec:SWform}) are used to probe the PDE and provide a source of {\em virtual} data \cite{kaltenbach2020incorporating}, i.e. the actual PDE never needs to be solved (section \ref{infusingVirtualObservables}).
 The formulation enables us to recast the problem as one of probabilistic inference (section \ref{sub:probInference}). We show that selecting at random a few of those residuals at each iteration is sufficient to learn a probabilistic surrogate. 
 While, as we discuss, various existing DNN-based architectures can be readily integrated into this framework, 
 we advocate instead a physics-aware implicit solver (section \ref{sec:ApproximatingDensity}) that consists of a (much) coarser, discretized version of the original PDE, which provides the requisite information bottleneck for high-dimensional problems and enables generalization in out-of-distribution settings (e.g. different boundary conditions).
We extend the framework to multiscale problems by proposing mPANIS (section \ref{sec:CGLearning}) and show that a surrogate can be learned for the effective (homogenized) solution without ever solving the reference problem \cite{van_bavel_efficient_2023}.  
We further demonstrate how the proposed framework can accommodate and generalize several existing learning objectives and architectures while yielding probabilistic surrogates that can quantify predictive uncertainty
(sections \ref{sec:panispred} \ref{sec:mpanispred}).
We substantiate its capability in the context of random heterogeneous materials where the input parameters represent the material microstructure (section \ref{sec:numerical}). Apart from comparative estimates with Physics Informed Fourier Neural Operators (PINO) \cite{li2021physics} and assessing performance with respect to the dimension of the parametric input vector, we pay special attention to problems that involve predictions under {\em extrapolative settings} i.e. different microstructures or boundary conditions as compared to the ones used in training. In such problems, and despite being extremely lightweight (e.g. it has up to three-orders of magnitude fewer training parameters as compared to PINO), we show that (m)PANIS can produce very accurate predictive estimates.

\section{Methodology}
\label{sec:Methodology}

We are concerned with parametric PDEs which can be generally written as:

\be
\begin{array}{ll}
\mathcal{L}(c(\bs{s};~\bx), u(\bs{s}))=0,~~~ \bs{s} \in \Omega & \textrm{((non)linear differential operator)} \\
\mathcal{B}(c(\bs{s};~\bx), u(\bs{s}))=0, \quad \bs{s} \in \pa \Omega & \textrm{(boundary-conditions  operator)} \\
\end{array}
\ee
where $\bs{s}$ denotes space, $\Omega \subset \RR^{d_s}$ the problem domain,  $c(\bs{s};~\bx)$ denotes an input function parametrized  by $\bx \in \mathcal{X} \subset \RR^{d_{\bx}}$ (e.g. microstructure or material property field) and  $u(\bs{s})$\footnote{The solution $u(\bs{s})$ implicitly depends on the parametric input $\bx$ which we omit in order to simplify the  notation.} the sought solution to the boundary value problem above.

In the following, we adopt a discretized representation of the solution, i.e. $u(\bs{s})\xrightarrow[]{\text{discretize}} \by$ where $\by \in \mathcal{Y} \subset \RR^{d_{\by}}$. We assume that the adopted discretization is fine enough to provide an acceptable approximation to the actual solution, which, in general, implies that $dim(\by)>>1$. The vector $\by$ could represent coefficients with respect to a given set of functions $\{\eta_i(\bs{s})\}_i^{d_{\by}}$ 
(e.g. FE shape functions, Radial Basis Functions,  Fourier, Wavelet):
\begin{equation}
    u_{\by}(\bs{s}) = \sum_{i=1}^{d_{\by}} y_i \eta_i(\bs{s}), ~~~ i=1,2,3,...,d_{\by}
    \label{eq:trialSolRepresentation}
\end{equation}
or the weight/biases of a  neural network.  While emphasis has been directed recently on neural operators that approximate maps between infinite-dimensional,  Banach spaces, we note that as demonstrated in \cite{fanaskov_spectral_2023}, spectral neural operators can be developed by employing truncated (i.e. finite-dimensional)   Chebyshev or Fourier series (i.e. when $\eta_i(\bs{s})$ in \refeq{eq:trialSolRepresentation} take the form of Chebyshev polynomials or sine/cosines) which can be superior in overcoming systematic bias caused by aliasing errors.

We propose Physics-Aware Neural Implicit Solvers (PANIS), a framework that casts the problem of developing surrogates for parametrized PDEs as one of probabilistic inference and which relies on the use of:
\bi
\item weighted residuals as virtual data (i.e. we never use {\em labeled} data in the form of solution pairs $(\bx,\by)$) which are ingested probabilistically with the help of a virtual likelihood, and
\item  an implicit solver at the core of the probabilistic surrogate constructed which provides the requisite information bottleneck in order to deal with high-dimensional problems (i.e. when $d_{\bx},d_{\by}>>1$) and to generalize in out-of-distribution settings.
\ei
The output of PANIS is  a {\em probabilistic} surrogate, which will be expressed in the form of a density $q(\by |\bx)$ that can predict the (discretized) solution for any input $\bx$. 
We note that the predictive uncertainty in this case is not due to aleatoric stochasticity in the governing equations (although such cases can be accommodated as well) but of epistemic nature due to the use of partial, and generally small, data in constructing this surrogate as explained in the sequel. 

\subsection{Strong and Weak form  of a boundary-value problem}
\label{sec:SWform}

We view the governing PDE as an infinite source of information, which we probe with the help of {\em weighted residuals}. These constitute the (virtual) data with which we attempt to train our surrogate.
In order to illustrate this concretely and clearly, we use as an example a linear, elliptic PDE. The methodological elements can then be generalized to other types of PDEs.
In particular, we consider the boundary value problem:

\be
\begin{array}{rl}
 \nabla (-c(\bs{s};~ \bx) \nabla u(\bs{s}))=g, & \bs{s} \in \Omega\\
 u(\bs{s})=u_0, &   \bs{s} \in \Gamma_u \\
 -c(\bs{s};~ \bx) \nabla u(\bs{s}) \bs{n}=q_0, &  \bs{s} \in \Gamma_q =\pa \Omega - \Gamma_u
\end{array}
\label{eq:DarcyFlowEquation}
\ee

where $c(\bs{s};~ \bx)$ is a (random) conductivity/permeability/diffusivity field which depends on the parameters $\bx$. 
We denote with $\Gamma_u$ and $\Gamma_q$ the parts of the boundary  where Dirichlet and Neumann conditions are respectively defined. The symbol $\bs{n}$ indicates the outward, unit normal vector while the $u_0$ and $q_0$ denote the prescribed values.

We employ candidate solutions $u_{\by}$ which are represented by the finite-dimensional  $\by \in \RR^{d_{\by}}$ as discussed above and which are assumed to a priori satisfy the Dirichlet BCs \footnote{This requirement is consistent with traditional derivations but can be readily relaxed or removed in our formulation.}. These are combined with weighting functions $w(\bs{s}) \in \mathcal{W}$ which are assumed to be zero at the Dirichlet boundary i.e. $w(\bs{s})|_{\bs{s} \in \Gamma_u}=0$.

By employing integration-by-parts \cite{finlayson_method_1972} and for each such $w(\bs{s})$ we obtain the weighted residual:
\be
r_w(\by,\bx) = 0= \int_{\Gamma_q} w~ q_0 ~d\Gamma+\int_{\Omega} \nabla w~ c(\bs{s};~ \bx) ~\nabla u_{\by}~ d\bs{s} -\int_{\Omega} w~ g ~d\bs{s}.
\label{eq:weak1}
\ee

\noindent We note that depending on the choice of the weight functions $w$ (at least) six methods (i.e. collocation, sub-domain, least-squares, (Petrov)-Galerkin, moments) arise as  special cases \cite{finlayson_method_1972}. As it will become apparent in the sequel, the weighted residuals are used as data sources and not in  the traditional sense, i.e. to derive a discretized system of equations  for the approximate solution of the problem. Hence, we would be at liberty to consider alternate or even non-symmetric versions with respect to $u_{\bs{y}}$ and $w$ as for example expressions with  lower-order derivatives of the candidate solution $u_{\by}$, which are obtained  by applying  further integration-by-parts \cite{kharazmi_hp-vpinns_2021}.

\subsection{ Virtual Observables and Virtual Likelihhod} \label{infusingVirtualObservables}

Rather than solving the governing equations multiple times for different values of $\bx$ in order to obtain a {\em labeled} training dataset (i.e. pairs of $(\bx,\by)$) 
we treat weighted residuals as in \refeq{eq:weak1} as {\em virtual observables} \cite{kaltenbach2020incorporating}.
In particular, given   $N$ distinct weighting function $w_j$ and the corresponding residuals $r_{w_j}(\by,\bx)$, we assume that their values $\hat{r}_j$ have been {\em virtually} observed and are equal to $0$. 
The {\em virtual} observables $\hat{\bs{R}}=\{ \hat{r}_j=0 \}_{j=1}^N$ imply a {\em virtual} likelihood which is assumed to be of the form:

\be
\begin{array}{ll}
 p(\hat{\bs{R}} | \by, \bx) & =\prod_{j=1}^N p(\hat{r}_j=0 | \by, \bx) \\
 & = \prod_{j=1}^N \frac{\lambda}{2}  e^{ - \lambda |r_{w_j}(\by,\bx)|}.
 \end{array}
\label{eq:virtuallike}
\ee
The role of the hyper-parameter $\lambda$ is significant as it controls how the likelihood of a pair $(\bx,\by)$ will decay if the corresponding residuals $r_{w_j}(\by, \bx)$ are different from zero. A reasonable guideline is to set $\lambda^{-1}$ equal to the  tolerance value selected in a deterministic iterative solver. We note that alternative forms of the virtual likelihood are also possible (e.g. Gaussian), but the form above was found to yield better results in the numerical experiments performed.  The role of the virtual observables/likelihood is to  provide information from the governing equations {\em without} having to solve them. We discuss in the sequel how this virtual likelihood can be used to learn a surrogate and how this can be carried out by evaluating a few residuals at a time (i.e. without ever solving the weak form).

\subsection{Learning surrogates as probabilistic inference}
\label{sub:probInference}

Let $p(\bx)$ be a density on the parametric input. In the case of a random input, this is directly furnished by the problem definition. For deterministic problems, this could simply be a uniform density over the domain of possible values of $\bx$. 
We complement this with a {\em prior} density $p(\by|\bx)$. In the ensuing numerical experiments, a vague, uninformative such prior was selected. We note that its role is not to provide a good approximation of the sought output $\by$ given an input $\bx$ as this is to be achieved by  the {\em posterior} discussed later.

The combination of the aforementioned prior densities with the virtual likelihood of \refeq{eq:virtuallike} in the context of  Bayes' rule leads to the {\em joint} posterior $p(\bx, \by | \hat{\bs{R}})$ as follows:

\be
\begin{array}{ll}
p(\bx, \by | \hat{\bs{R}}) & =\cfrac{ p( \hat{\bs{R}} | \bx, \by) p(\by |\bx) p(\bx) }{ p(\hat{\bs{R}}) }. \\
\end{array}
\label{eq:postint}
\ee
We  emphasize that the posterior above is defined on the {\em joint} space of inputs $\bx$ and outputs $\by$. Apart from the effect of the prior, it assigns the highest probability to pairs of $\bx$ and $\by$ that achieve  $0$ residuals as is the case when they are a solution pair of the boundary value problem.

Since the posterior is generally intractable, we advocate a Variational-Inference scheme \cite{paisley2012variational}, 
 i.e. we seek an approximation from a parameterized family of densities $q_{\bpsi}(\bx,\by)$ by minimizing the Kullback-Leibler divergence with the exact posterior above. 
 If $\bpsi$ denotes the tunable parameters, this is equivalent  to maximizing the Evidence Lower BOund (ELBO) $\mathcal{F}(\bpsi)$ to the log-evidence $\log p(\hat{\bs{R}})$ \cite{blei2017variational}, i.e.:

\be
\begin{array}{ll}
 \log p( \hat{\bs{R}}) & = \log \int   p( \hat{\bs{R}} | \bx, \by) p(\by |\bx) p(\bx) ~d\by ~d\bx \\
 & \ge \left< \log \cfrac{   p( \hat{\bs{R}} | \bx, \by) p(\by |\bx) p(\bx) }{q_{\bpsi}(\bx,\by)} \right>_{q_{\bpsi}(\bx,\by)} \\
 & =\mathcal{F} (\bpsi), 
 \end{array}
 \label{eq:logevi}
\ee
where the brackets $<.>_{q}$ indicate an expectation with respect to $q$. 
Substitution of \refeq{eq:virtuallike} 
 in the expression above leads to:

\be
\begin{array}{ll}
 \mathcal{F} (\bpsi) &  = N\log \frac{\lambda}{2} - \lambda \sum_{j=1}^N \left< |r_{w_j}(\by,\bx)|\right> _{q_{\bpsi}(\bx,\by)}+ \left< \log \cfrac{p(\by |\bx) p(\bx) }{q_{\bpsi}(\bx,\by)}\right>_{q_{\bpsi}(\bx,\by)} \\
& = N\log \frac{\lambda}{2} - \lambda \sum_{j=1}^N \left< |r_{w_j}(\by,\bx)|\right> _{q_{\bpsi}(\bx,\by)}-KL\left(q_{\bpsi}(\bx,\by) ||  p(\by |\bx) p(\bx) \right). \\
 \end{array}
\label{eq:elbon}
\ee
Apart from the first term, which is irrelevant when $\lambda$ is fixed, we note that the objective above promotes $q_{\bpsi}$ that   minimize (on average) the absolute values of the $N$ residuals considered while simultaneously minimizing the KL-divergence from the prior density adopted. 

\noindent \textbf{Remarks:}\\
\bi
\item The KL-divergence term in \refeq{eq:elbon} can be thought of  as a regularization or penalty term, which  nevertheless arises naturally in the Bayesian framework adopted.  In other physics-informed schemes that have been proposed \cite{li2021physics, de2024error,kharazmi_hp-vpinns_2021}, ad-hoc parameters are generally selected as relative weights of these two terms \cite{krishnapriyan2021characterizing}. Similarly, alternative forms of virtual likelihoods  can give rise to the exact expressions found in competitive schemes. E.g. the use of a Gaussian likelihood would yield the sum of the {\em squares} of the $N$ residuals in the ELBO.
\item Furthermore such schemes generally employ collocation-type residuals in the physics-informed term \cite{raissi2019physics,wang2021learning,li2021physics}, which in our formulation arise as a special case  when the weight functions considered take the form of Dirac-deltas i.e. $w_j(\bs{s})=\delta(\bs{s}-\bs{s_j})$ where $\bs{s}_j$ correspond to the collocation points.
\item More importantly, by finding the optimal $q_{\bpsi}$ we obtain a {\em probabilistic surrogate} that is capable of providing probabilistic predictions of the solution $\by$ for any input $\bx$. Since the form of $q_{\bpsi}$ is critical to the accuracy of the proposed scheme we defer detailed discussions to Section \ref{sec:ApproximatingDensity}. As a final note we mention that if a degenerate $q_{\bpsi}$ is selected in the form of a Dirac-delta, then one obtains a {\em deterministic} surrogate as is the majority of the ones proposed in the literature \cite{nabian2018deep,lu2019deeponet,yu2018deep}. 
\ei

The evaluation and, more importantly,  the maximization of the ELBO $\mathcal{F}(\bpsi)$ requires the computation of the expectations with respect to $q_{\bpsi}$ of the absolute values of the $N$ weighted-residuals appearing in the virtual likelihood. We note that the evaluation of a weighted residual involves the computation of an integral over the problem domain (see e.g. \refeq{eq:weak1}) but not the solution of the PDE. These integrations can be carried out using deterministic or (quasi) Monte-Carlo-based numerical integration schemes \cite{morokoff1995quasi,zang2020weak}. Furthermore, if the weight functions have support on a subset of the problem domain $\Omega$, additional efficiencies can be achieved in the numerical integration.

Independently of the particulars, it is readily understood that as $N$ increases, so does the information we extract from the governing PDE, but so does also the computational effort. 
In order to improve the computational efficiency of these updates, we propose a Monte Carlo approximation of the pertinent term in the ELBO in \refeq{eq:elbon}. In particular

\begin{equation}
\sum_{j=1}^N \left< |r_{w_j}(\by,\bx)| \right> _{q_{\bpsi}(\bx,\by)} \approx \frac{N}{M} \sum_{m=1}^M \left< |r_{w_{j_m}}(\by,\bx)| \right> _{q_{\bpsi}(\bx,\by)}, ~\text{where } j_m \sim Cat\left(N,\frac{1}{N}\right).
\label{eq:randres}
\end{equation}
In essence, we sample at random and with equal probability $M<N$ weight functions out of the $N$ and use them to approximate the sum. The specifics for the set of $N$ weight functions, which is subsampled, are described in subsection \ref{sub:2D_Darcy}. As $M$ increases, so does the Monte Carlo error decrease, but so does the computational effort increase. Since the Monte Carlo estimator is unbiased even for $M=1$,  evaluating a single residual is sufficient to guarantee convergence (see Algorithm \ref{alg:RandomResidual}).

   In combination with \refeq{eq:elbon} (we omit the first term, which is independent of $\bpsi)$, this yields the following Monte Carlo approximation to the ELBO:

\be
\begin{array}{ll}
 \mathcal{F} (\bpsi) &  \approx  -\lambda \frac{N}{M} \sum_{m=1}^M \left< |r_{w_{j_m}}(\by,\bx)| \right> _{q_{\bpsi}(\bx,\by)} + \left< \log \cfrac{p(\by |\bx) p(\bx) }{q_{\bpsi}(\bx,\by)}\right>_{q_{\bpsi}(\bx,\by)} ~\text{where } j_m \sim Cat\left(N,\frac{1}{N}\right).

\end{array}
\label{eq:elboRandomResidual}
\ee
We discuss in subsection \ref{sec:ApproximatingDensity} and in more detail in \ref{appendix:elboTerms} how gradients of $\mathcal{F}$ can be computed and used in order to carry out the ELBO maximization. We also note that if a parameterized prior, say  $p_{\bt}(\by|\bx)$ were to be adopted, then the optimal $\bt$ could also be identified by maximizing the corresponding ELBO \cite{rixner2021probabilistic}. While such structured or informative priors could be highly beneficial in accelerating inference and improving predictive accuracy, they are not explored in this paper. 

Finally, we note that even though the method described above for subsampling weighting functions and residuals  is simple and efficient, other more intelligent procedures could lead to weighting functions of superior informational content.
One such idea would be to start with a null set of residuals (i.e. the prior) and successively add one  residual at a time (i.e. selecting a new $w$) based e.g. on maximizing the KL-divergence between the current posterior (or its approximation) and the posterior with the additional  residual. 

Another scheme could involve the parameterization of $w$ (e.g. by some parameters $\bphi$) and, subsequently, the iterative maximization of the ELBO w.r.t the parameters $\bpsi$ followed by a minimization of the ELBO w.r.t. the parameters $\bphi$ for every optimization cycle (i.e. a saddle-point search).
We defer a more thorough investigation of these possibilities as well as their effect in reducing the computational effort for future work.

\subsection{Form of the Approximating Density $q_{\bpsi}$ and Implicit Solvers}
\label{sec:ApproximatingDensity}

In selecting the form of $q_{\bpsi}$ and without loss of generality  we postulate: 
\be
q_{\bpsi}(\bx,\by) = q_{\bpsi}(\by|\bx) q(\bx) = q_{\bpsi}(\by|\bx) p(\bx) 
\label{eq:AppoxPostForm}
\ee 
i.e. we use the given density of the parametric input and attempt to approximate the input-to-output map with $q_{\bpsi}(\by|\bx)$ where $\bpsi$ denotes the tunable parameters. We note that alternate decompositions might be advisable in the context of inverse problems but are not discussed in this work.
We note also  that a deterministic formulation can be readily adopted by selecting a degenerate density in the form of a Dirac-delta, i.e.:
\be
q_{\bpsi}(\by|\bx)=\delta(\by-\bs{\mu}_{\bpsi} (\bx)).
\ee
The sought function $\bs{\mu}_{\bpsi} (\bx)$ in this case can take the form of {\em any}
 of the proposed NN-based architectures such as Fully Connected Neural Networks (FCNNs), Convolutional Neural Networks (CNNs) \cite{zhu2019physics}, Deep Neural Operators (DeepONets) \cite{lu2019deeponet}, Graph Neural Operators (GNOs) \cite{li2020neural} or Fourier Neural Operators (FNOs) \cite{kovachki2021neural}, in which case $\bpsi$ would correspond to the associated NN-parameters.

While all these architectures have produced good results in a variety of problems, it is easily understood that as the variability and the dimension of the input $\bx$ (as well as of the sought output $\by$) increase, so must the complexity of the aforementioned NN-based schemes increase. At a practical level, it is not uncommon that one must train for millions or billions of parameters $\bpsi$ (\cite{zhu2019physics, lu2019deeponet, wang2021learning}). This situation is much more pronounced in {\em multiscale} problems which are discussed in Section \ref{sec:CGLearning} where input and output de facto imply very high-dimensional representations.

More importantly, as such interpolators are agnostic of the physical context, they face significant problems in generalizing \cite{lu2019deeponet, li2020fourier}. For example, if out-of-distribution predictions are sought for a significantly different input $\bx$ as compared to the ones used in training, then the predictive accuracy can significantly drop \cite{wang2021learning}. In addition, and in the context of PDEs in particular, if one were to seek predictions under a different set of boundary conditions than the ones employed during training, most likely, these predictions would be significantly off and by a large margin (see Section \ref{sec:numerical}). 

One could claim that enlarged datasets, bigger models, or progress in pertinent hardware could alleviate or overcome these difficulties. Nevertheless, none of these strategies addresses the fundamental, underlying question which is how to find the requisite information bottlenecks that lead to  drastic dimensionality reductions and lightweight surrogates
 and how to achieve this in a problem-aware manner that is capable of producing accurate  predictions under extrapolative settings \cite{grigo_physics-aware_2019}.

In view of these objectives and following previous work \cite{grigo_physics-aware_2019,rixner2021probabilistic}, we propose incorporating coarse-grained  (CG) versions of the governing PDE as an implicit solver \cite{belbute2020combining,um2020solver} in the surrogate constructed. 
We elaborate on this in the subsequent sections, and more importantly, we extend it to {\em multiscale} problems, which pose a unique set of challenges. 

For the purposes of the Variational Inference proposed, we employ a multivariate Gaussian of the form:
\be
q_{\bpsi}(\by|\bx)  = \mathcal{N}(\by~|~\bs{\mu}_{\bpsi} (\bx), \boldsymbol{\Sigma}_{\bpsi})
\label{eq:FormofTheAproxPosterior}
\ee
the mean and covariance of which are discussed in the sequel.

\begin{figure}[!t]
  \centering
  \includegraphics[width=0.8\textwidth]{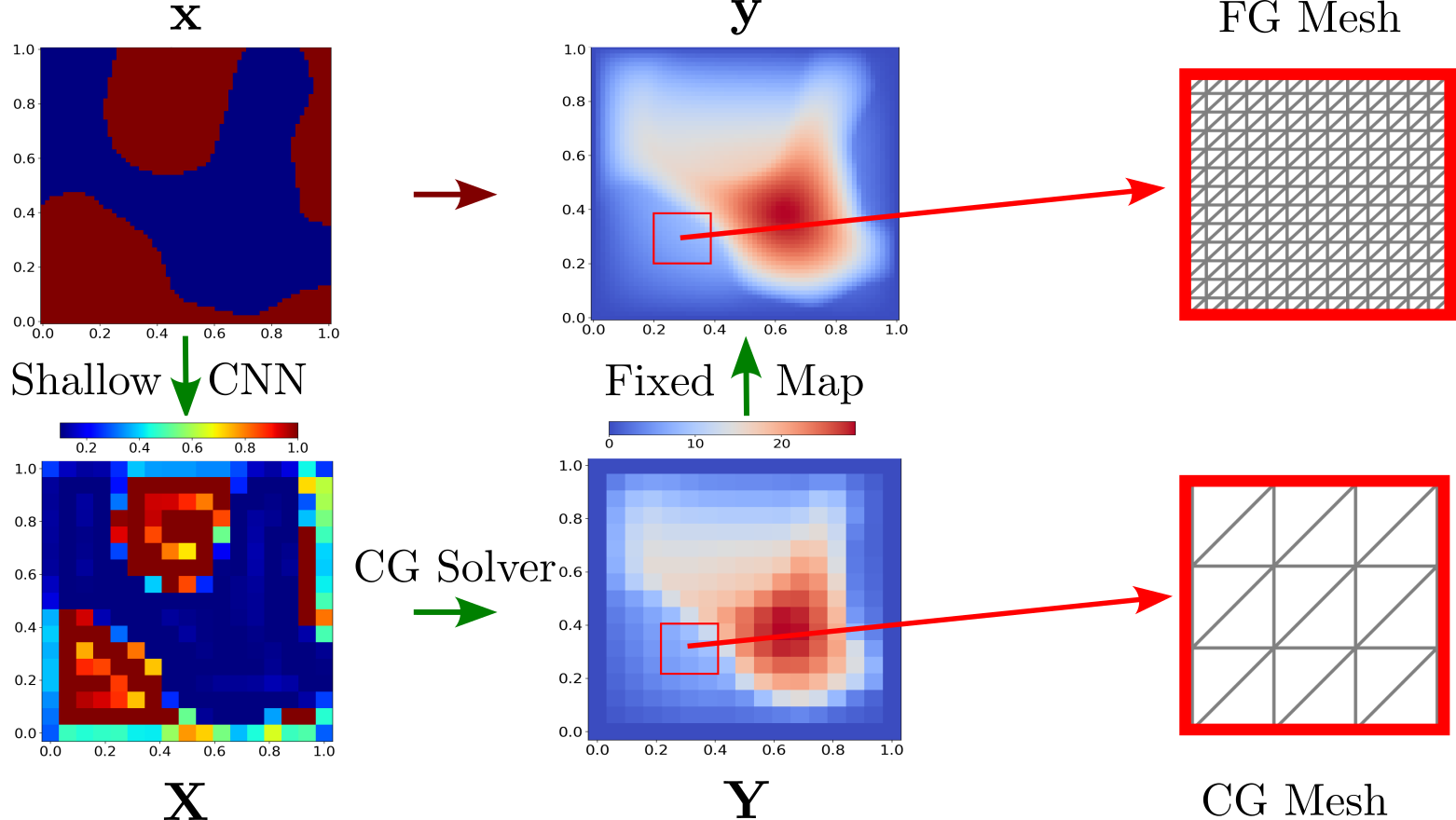}
  \caption{Schematic illustration of the proposed, physics-aware architecture for parametrizing the mean solution $\bs{\mu}_{\bpsi} (\bx)$ as described in subsection \ref{sec:ApproximatingDensity}. 
  }
  \label{fig:CGPhysicsArchitecture}
\end{figure}

In the context of  materials problems and as in the model formulation  of \refeq{eq:DarcyFlowEquation} where the inputs $\bx$ parametrize a, potentially multiscale, material property field,  we consider a (much) coarser, discretized version (see e.g. Figure \ref{fig:CGPhysicsArchitecture}) of the governing PDE where $\bxx, \byy$ denote the respective, discretized input and output. 
By construction, their dimension is much lower as compared to  their reference, fine-grained counterparts $\bx,\by$. As such, the solution of the associated (non)linear algebraic equations to obtain the solution $\byy$ for any input $\bxx$  is much faster.
As we have shown in \cite{grigo_physics-aware_2019}, the $\bxx$ could be thought of as some effective material properties which serve as the requisite information bottleneck to the  problem. Furthermore, it is possible that  a different (discretized) PDE  could be used as a stencil for the coarse-grained model.

We complement the implicit, differentiable, and vectorized, input-output map  $\byy(\bxx)$  of the CG model, with two additional components:
\bi
\item a \textbf{fine-to-coarse input map} denoted by $\bxx_{\bpsi_x}(\bx)$ that we express with a shallow CNN parameterized by $\bpsi$. This consists merely of 2 to 3 convolution and deconvolution layers, as well as a few, average pooling layers. To improve stability we employ batch normalization wherever is required, while the only kind of activation function used was Softplus. The number of feature maps of each layer is limited, and the number of total trainable parameters is {\em three orders of magnitude lower} compared to other state-of-the-art models used on the same problems (more details in \ref{appendix:shallowCNN}).

\item a \textbf{coarse-to-fine output map} denoted by $\by(\byy)$ that  attempts to reconstruct the reference, fine-grained (FG) output $\by$ given the CG output $\byy$. In our illustrations, this map was fixed (i.e. it had no trainable parameters), which has the desirable effect of forcing $\bxx$ to attain physically meaningful values in the context of materials problems. In particular a {\em linear map} of the form $\by=\bs{A} \byy$ was used where the matrix $\bs{A}$ was pre-computed so as to minimize the mean-square error in the resulting solution fields (more details in \ref{appendix:ytoYTransform}).  A very important benefit of the use of this implicit solver and of the associated map is the direct imposition of boundary conditions on the CG model, i.e. on $\byy$, which are automatically transferred to $\by$. In this manner and we demonstrate in the numerical illustrations, our model can generalize under varying boundary conditions.

\ei
The composition of these three functions leads to:
\be
\bs{\mu}_{\bpsi} (\bx)=\bs{A} ~ \byy\left(\bxx_{\bpsi_x}(\bx) \right).
\label{eq:formOfmeanPrediction}
\ee
In the formulation employed in the ensuing numerical experiments, the sole trainable parameters are associated with the inner layer i.e. the fine-to-coarse map from $\bx$ to $\bxx$. More complex formulations involving trainable parameters in the other two layers could be readily accommodated.
We note that all three layers are differentiable, and each evaluation requires the solution of the CG model. Differentiation of $\bs{\mu}_{\bpsi}$ with respect to $\bpsi_x$ as needed for training requires a backward pass which can be computed with Automatic Differentiation \cite{bartholomew2000automatic, baydin2018automatic} and/or adjoint formulations of the governing equations of the CG model \cite{jameson2003aerodynamic, papoutsis2016continuous}.

With regards to the covariance matrix $\boldsymbol{\Sigma}_{\bpsi}$  in Equation \ref{eq:FormofTheAproxPosterior} and given the high dimension of $\by$, we advocate a low-rank approximation of the form:

\begin{equation}
\label{eq:CovMatrix1}
\boldsymbol{\Sigma}_{\bpsi} = \bs{L}_{y}\bs{L}_{y}^T+\sigma_{y}^2 \mathbf{I}_{d_{\by}},
\end{equation}

\noindent where $\bs{L}_{y}$ is a rectangular  matrix  of dimension  $d_{\bs{y}} \times d_{\by}'$ (with $d_{\by}'<< d_{\by}$) 
and $\sigma_{\bpsi}^2 \in \RR_+$ is a scalar.  
The first term is responsible for introducing correlation between the solution vector's entries whereas the second term accounts for the residual,  global uncertainty in the solution. 
The low-rank approximation ensures that the number of unknown parameters in $\bs{L}_y$ grows {\em linearly} with the dimension of $\by$.

In summary, the vector of trainable parameters $\bpsi$ consists of:
\be
\bpsi=\left\{\bpsi_x, \bs{L}_y, \sigma^2_y \right\}.
\ee

\noindent Maximization of the ELBO $\mathcal{F}$ with respect to $\bpsi$ is carried out in the context of Stochastic Variational Inference (SVI) \cite{hoffman2013stochastic,paisley2012variational}. This entails Monte Carlo approximations for the intractable expectations appearing in the gradient of $\nabla_{\bpsi} \mathcal{F}$ and Stochastic Gradient Ascent (SGA). In our experiments, we used ADAM for the SGA because of its adaptive learning rates, momentum-based updates, and its effectiveness as an optimization algorithm \cite{kingma2014adam, tian2023recent}.
Critical to achieving Monte Carlo estimators with low error is the use of the reparametrization trick \cite{vatanen2013pushing, kingma2015variational}, which for $\by$ sampled from $q(\by|\bx)$ implies:

\be
\by = \bs{\mu}_{\bpsi_x} (\bx) + \bs{L}_{y} \bs{\varepsilon_1} + \sigma_{y} \bs{\varepsilon_2}, ~~~ \bs{\varepsilon_1} \sim \mathcal{N}\left( \bs{0}, ~ \bs{I}_{d_{\by}'} \right),  ~ \bs{\varepsilon_2} \sim \mathcal{N}\left( \bs{0}, ~ \bs{I}_{d_{\by}} \right),
\label{eq:yreparam}
\ee

\noindent where $dim(\bs{\varepsilon_1})=d_{\by}'$, $ dim(\bs{\varepsilon_2})=d_{\by}$.
In Algorithm \ref{alg:RandomResidual} we summarize the main algorithmic steps.

\begin{algorithm}[!t]
\begin{algorithmic}[1]
    \State  Select $\lambda$, $N$, $M$; Initialize $\bpsi\gets \boldsymbol{\psi_0}$;
    \State Generate a set of $N$ weighting functions \Comment{see subsection \ref{sub:probInference}}
    \State $\ell \gets 0$ \Comment{Initialize iteration counter}
    \While{$\mathcal{F}_{\bpsi}$ not converged} 
        \State Draw $M$ weight functions at random out of the $N$ \Comment{see \refeq{eq:randres}}
        \State Draw $\bx \sim p(\bx)$ 
        \State  Generate $\by \gets \bs{\mu}_{\bpsi_x} (\bx) + \bs{L}_{y} \bs{\varepsilon_1} + \sigma_{y} \bs{\varepsilon_2}, ~~~ \bs{\varepsilon_1} \sim \mathcal{N}\left( \bs{0}, ~ \bs{I}_{d_{\by}'} \right),  ~ \bs{\varepsilon_2} \sim \mathcal{N}\left( \bs{0}, ~ \bs{I}_{d_{\by}} \right)$ 
        \State Approximate $\mathcal{F}_{\bpsi}$ \Comment{\textbf{Monte Carlo} Approximation according to Equation \eqref{eq:elboRandomResidual}}
        \State Estimate $\nabla_{\bpsi} \mathcal{F}_{\bpsi}$ \Comment{\textbf{Backpropagation} for estimating $\nabla_{\bpsi} \mathcal{F}_{\bpsi}$}
        \State $\bpsi_{\ell+1} \gets \bpsi_{\ell} + \boldsymbol{\rho}^{(\ell)} \odot \nabla_{\bpsi} \mathcal{F}_{\bpsi}$ \Comment{\textbf{SGA} using \textbf{ADAM} for determining the learning rate $\boldsymbol{\rho}^{(\ell)}$}
        \State $\ell \gets \ell + 1$
    \EndWhile
\end{algorithmic}
\caption{PANIS Training Algorithm}
\label{alg:RandomResidual}
\end{algorithm}

\subsection{PANIS - Posterior Predictive Estimates}
\label{sec:panispred}
Upon convergence of the aforementioned SVI scheme and the identification of the optimal $\bpsi$ parameters, the approximate posterior $q_{\bpsi}(\by|\bx)$ in \refeq{eq:FormofTheAproxPosterior} 
 can be used for posterior predictive estimates. For each test input $\bx$, samples from $q_{\bpsi}(\by|\bx)$ can readily be drawn as in \refeq{eq:yreparam}.
We note that for each such sample, the solution of the CG model (i.e. the computation of $\byy$ for the corresponding $\bxx_{\bpsi_x}(\bx)$)
 would be required. 
For the  representation of $u_{\by}$ in Equation \eqref{eq:trialSolRepresentation} (which is linear with respect to $\by$) posterior predictive estimates of the continuous PDE-solution $u_{\by}(\bs{s})$ can be readily obtained as detailed in Algorithm \ref{alg:MakingPredictions} by making use of $\bs{\mu}_{\bpsi}(\bx)$ in \eqref{eq:formOfmeanPrediction} and $\bs{\Sigma}_{\bpsi}$ in \eqref{eq:CovMatrix1}. 

In order to comparatively assess the performance of the proposed framework, for each of the problems in the numerical illustrations  we consider a validation dataset $\mathcal{D}_v = \{ \mathbf{x}^{(j)}, \bs{u}^{(j)}\}_{j=1}^{N_{v}}$ consisting of $N_{v}$ input-solution pairs, which were obtained by using a conventional FEM solver on a fine mesh. Each vector $\bs{u}^{(j)}$ refers to the discretized version of $u^{(j)}(\bs{s})$, the specifics of which are described in Section \ref{sec:numerical}. The following metrics were computed:

\textbf{Coefficient of Determination $R^2$:}
The coefficient of determination is a widely used metric \cite{coeffDetermZhang} to quantify the accuracy of point estimates, and it is defined as follows

\begin{equation}
R^2 = 1 - \frac{\sum_{j=1}^{N_j} \| \bs{u}^{(j)}- \bs{u}^{(j)}_{\mu} \|_2^2}{\sum_{j=1}^{N_j} \| \bs{u}^{(j)} - \bar{\bs{u}}^{(j)} \|_2^2},
\label{eq:RSquared}
\end{equation}

where $\bs{u}^{(j)}_{\mu}$ is the mean discretized solution field as predicted from the predictive posterior of our trained model for each  $\bx^{(j)}$ in $\mathcal{D}_v$,  $\bar{\bs{u}}^{(j)} = \frac{1}{N_j} \sum_{j=1}^{N_j} \bs{u}^{(j)}$ is the sample mean over the validation dataset and $||.||_2$ denotes the $L_2-$norm over the problem domain $\Omega$. The maximum score is $R^2=1$, which is attained when the predictive mean estimates coincide with the true solutions. It is noted that negative $R^2$ values are also valid since the second term is merely weighted (not bounded) by the empirical variance of the validation dataset.

\textbf{Relative $L_2$ Error $\epsilon$:} The relative $L_2$ error is commonly used as an evaluation metric for comparisons between state-of-the-art surrogate models for stochastic PDEs \cite{zhu2019physics, lu2022comprehensive, wang2021learning}. It is defined as:

\begin{equation}
\epsilon = \frac{1}{N_j} \sum_{j=1}^{N_j}\frac{ \| \bs{u}^{(j)} - \bs{u}^{(j)}_{\mu} \|_2}{\| \bs{u}^{(j)} \|_2}.
\label{eq:relError}
\end{equation}

The overwhelming majority of competitive methods produce deterministic predictions, i.e. point estimates, which prevent a direct comparison with the predictive posterior $q_{\bpsi}(\by|\bx)$ obtained by our framework. Nevertheless, in the illustrations of Section \ref{sec:numerical}, we report upper/lower posterior uncertainty bounds obtained as detailed in Algorithm \ref{alg:MakingPredictions}.

\begin{algorithm}[t]

\begin{algorithmic}[1]
    \State Compute  {$u_{\mu}(\bs{s}) \leftarrow \sum_{i=1}^{N_{\eta}} \mu_{\bpsi} (\bx)_i \eta_i(\bs{s})$} \Comment{Posterior Mean of Solution Field };
    \State Compute  {$u_{\sigma}(\bs{s}) \leftarrow \sqrt{ \sum_{i,j=1}^{N_{\eta}}  \eta_i(\bs{s}) \bs{\Sigma}_{\bpsi, ij} \eta_j(\bs{s}) }$} \Comment{Posterior St. Deviation of Solution Field}

    \State $u_{U}(\bs{s}) \leftarrow u_{\mu}(\bs{s}) + 2 u_{\sigma}(\bs{s})$ \Comment{Upper Uncertainty Bound Field}
    \State $u_{L}(\bs{s}) \leftarrow u_{\mu}(\bs{s}) - 2 u_{\sigma}(\bs{s})$ \Comment{Lower Uncertainty Bound Field}
\end{algorithmic}

\caption{Posterior Predictive Estimates for the solution (\refeq{eq:trialSolRepresentation}) for a test input $\bx$.}

\label{alg:MakingPredictions}

\end{algorithm}
\section{Learning solutions to multiscale PDEs}
\label{sec:CGLearning}

As mentioned in section \ref{introduction}, a challenging  case in which data-driven (physics-informed or not) methods struggle,  are multiscale problems. In the context of heterogeneous media 
 such problems naturally arise as the scale of variability of the underlying microstructure becomes smaller. Capturing this would, in general, require very high-dimensional input $\bx$ and output $\by$ vectors, and while existing neural architectures could be employed without alterations, they would require bigger and deeper nets (i.e. more trainable parameters) as well as, in general, more training data.   
 In multiscale PDEs, we  are interested in finding an effective or homogenized model that captures the macroscale components of the solution. Classical homogenization techniques rely on very special features such as periodicity at the fine-scale \cite{torquato2002random}. Several   numerical homogenization techniques
 \cite{weinan2011principles} have been developed  which  consistently upscale fine-scale information which, although small in magnitude, can have a significant effect in the macroscale solution \cite{pavliotis2008multiscale}. In the context of weighted residuals, setting these fine-scale components of the solution to zero would lead to significant errors in the coarse-scale components \cite{sanchez1987homogenization, geers2017homogenization}. 
 
While several data-driven strategies have been proposed for learning such effective models \cite{zhang_gfinns_2021,hernandez2022thermodynamics,masi2022multiscale,cueto2023thermodynamics}, to the best of our knowledge  all these rely on labeled data i.e. pairs of inputs $\bx$ and outputs/solutions  $\by$  of the multiscale PDE. In the following, we propose \textbf{ multiscale PANIS} (mPANIS), which is an extension of  the previously described framework that can  learn such effective models {\em without} ever solving the multiscale PDE but solely by computing weighted residuals (and their derivatives). The enabling component remains the physics-aware, implicit solver at the core of our surrogate, which can be learned by a small number of inputs and corresponding residuals.

\subsection{Model extension for multiscale PDEs (mPANIS)}

We additively decompose the solution-output $\by$ as:
\be 
\by=\by_c +\by_f,
\label{eq:ydeco}
\ee
 where $\by_c$ accounts for the coarse-scale features we are interested in and $\by_f$ for the fine-scale fluctuations that are not of interest.
In the context of the probabilistic models advocated, such a task would entail learning $q_{\bpsi}(\by_c|\bx)$, and  $q_{\bpsi}(\by_f|\bx)$.
On the basis of the previous single-scale model, we employ  a 
$q_{\bpsi}(\by_c|\bx)$ of the form:
\be
q_{\bpsi}(\by_c|\bx)=\mathcal{N} (\by_c | \bs{\mu}_{\bpsi}(\bx), \bs{\Sigma}_{\bpsi}),
\label{eq:qyc}
\ee
where as in \refeq{eq:formOfmeanPrediction} we make use of the implicit solver and the associated fine-to-coarse $\bxx_{\bpsi_x}(\bx)$ and coarse-to-fine $\bs{A} \byy$ (see section \ref{sec:ApproximatingDensity}) maps such that $\bs{\mu}_{\bpsi} (\bx)=\bs{A} ~ \byy\left(\bxx_{\bpsi_x}(\bx) \right)$.
Hence, the mean of $\by_c$ lives on the subspace spanned by the columns of the $\bs{A}$ matrix, and its reduced coordinates are determined by the output $\byy$ of the implicit solver.
We also introduce a learnable correlation on this subspace by adopting a $d_{\by} \times d_{\by}$ covariance matrix $\bs{\Sigma}_{\bpsi}=\bs{A} ~\bs{S}_{\bpsi}~\bs{A}^T$ where the inner $d_{\byy} \times d_{\byy}$ covariance matrix $\bs{S}_{\bpsi}$ is:
\be
\bs{S}_{\bpsi}=\bs{L}_Y \bs{L}_Y^T+\sigma_Y^2 \bs{I}_{d_{\byy}}.
\ee
The rectangular matrix $\bs{L}_Y$ of dimension $d_{\byy} \times d_{\byy}'$ (with $d_{\byy}' < d_{\byy}$) captures the principal directions where correlation appears and the scalar $\sigma_Y \in \RR_+$ the residual uncertainty. We emphasize that in this manner, the posterior uncertainty for $\by_c$ will be concentrated in the subspace spanned by the columns of the $\bs{A}$ matrix.

Given the aforementioned representation of $\by_c$, the fine-scale features $\by_f$ in \refeq{eq:ydeco} are expressed as:
\be
\by_f= \bs{A}_{\perp} \by_f',
\label{eq:yf}
\ee
where the columns of matrix $\bs{A}_{\perp}$ span the orthogonal complement of the subspace spanned by the columns of $\bs{A}$ above. This matrix can be readily computed, and its use ensures that the $\by_f$ complements $\by_c$ in representing the solution $\by$.
Given the very high dimension of $\by_f'$, which is commensurate with that of the full solution $\by$ and which would pose significant, if not insurmountable, challenges in learning $q_{\bpsi}(\by_f'|\bx)$, we propose the following form which is justified by the ability of the $\by_c$-part of the model to learn from very few inputs $\bx$.

In particular, consider an empirical approximation of the input density $p(\bx)$ consisting of $K$ samples, i.e.: 
\begin{equation}
p(\bx) = \frac{1}{K} \sum_{k=1}^K  \delta(\bx - \bx_k).
\label{eq:pemp}
\end{equation}

We propose employing an approximate posterior:
\be
q_{\bpsi}(\by_f' | \bx_k)= \delta(\by_f' - \by_{f,k}').
\label{eq:qyf}
\ee
This implies that for each atom $\bx_k$ in \refeq{eq:pemp}, we make use of a distinct Dirac-delta, centered at an (unknown) $\by_{f,k}'$. 
The corresponding $\{\by_{f,k}' \}_{k=1}^K$ become part of the vector of the unknown parameters $\bpsi$ which are determined by maximizing the ELBO. While the dimension of each $\by_{f,k}'$ can be high, the number $K$ of $\bx_k$-samples can be small due to the capacity of the proposed model architecture to learn from small data.
In summary, the vector of parameters $\bpsi$ that would need to be found by maximizing the corresponding ELBO (see subsequent section) in the multiscale version of our framework consists of:
\be
\bpsi=\left\{\bpsi_x, \bs{L}_Y, \sigma^2_Y, \{\by_{f,k}'\}_{k=1}^K~ \right\}.
\ee

\subsection{Algorithmic implementation}
\label{sec:algorithmicImplem}

Given the representation of the solution $\by$ as in \refeq{eq:ydeco}, the latent variables to be inferred in the Variational Inference framework advocated are now $\by_c$ and $\by_f'$ (\refeq{eq:yf}).
The associated weighted residuals, which are used as virtual observables, can be readily re-expressed with respect to $\by_c$ and $\by_f'$ as $\{r_{w_j}(\by_c,\by_f', \bx)\}_{j=1}^N$. In combination with priors $p(\by_c|\bx), p(\by_f'|\bx)$ \footnote{As in the single-scale case, uninformative priors were used here in  the form of $\mathcal{N}\left( \bs{0}, \sigma^2 \bs{I}\right)$, where $\sigma^2 = 10^{16}$}, we obtain analogously to \refeq{eq:elbon} the following ELBO\footnote{We omit any constant terms}:
\be
\mathcal{F}(\bpsi)=- \lambda \sum_{j=1}^N \left< |r_{w_j}(\by_c,\by_f', \bx)|\right> _{q_{\bpsi}(\bx,\by_c,\by_f')}+ \left< \log \cfrac{p(\by_c |\bx) p(\by_f'|\bx) p(\bx) }{q_{\bpsi}(\bx,\by_c,\by_f')}\right>_{q_{\bpsi}(\bx,\by_c,\by_f')}.
\label{eq:elboRandomResidualMultiscale}
\ee

The particular form of $q_{\bpsi}(\by_f'|\bx)$ in \refeq{eq:qyf} in combination with the empirical approximation of $p(\bx)$ in \refeq{eq:pemp} yield the following result for the first term in the ELBO:
\be
\sum_{j=1}^N \left< |r_{w_j}(\by_c,\by_f', \bx)|\right> _{q_{\bpsi}(\bx,\by_c,\by_f')} =\frac{1}{K} \sum_{k=1}^K \sum_{j=1}^N \left< |r_{w_j}(\by_c,\by_{f,k}', \bx_k)|\right> _{q_{\bpsi}(\by_c |\bx)}.
\label{eq:residualTermMulti}
\ee
This lends itself naturally to Monte Carlo approximations both in terms of the index $k$ (i.e. by subsampling only a subset of the $K$ atoms $\bx_k$ in \refeq{eq:pemp}) as well as in terms of the index $j$ i.e. over the considered residuals as explained in \refeq{eq:randres}. 
Expectations with respect to $q(\by_c|\bx)$ are also approximated by Monte Carlo in combination with the reparametrization trick \cite{vatanen2013pushing, kingma2015variational}, which according to \refeq{eq:qyc} can make use of $\by_c$ samples generated as follows:
\be
\bs{y}_c = \bs{A} \left( \byy\left(\bxx_{\bpsi_x}(\bx) +\bs{L}_Y \bs{\varepsilon}_1 +\sigma_Y \bs{\varepsilon}_2 \right) \right), \quad \bs{\varepsilon_1} \sim \mathcal{N}\left( \bs{0}, \bs{I}_{d_{\byy}'} \right), \quad \bs{\varepsilon_2} \sim \mathcal{N}\left( \bs{0}, \bs{I}_{d_{\byy}} \right),
\label{eq:ycreparam}
\ee
\noindent where $dim(\bs{\varepsilon_1})=d_{\byy}'$, $dim(\bs{\varepsilon_2})=d_{\byy}$.
Details about the ELBO and the computation of its gradient with respect to $\bpsi$ are contained in \ref{appendix:elboTerms}. In Algorithm \ref{alg:RandomResidualMulti}, we summarize the basic steps involved.

\noindent \textbf{Remarks:} \\
\bi
\item One of the common issues in pertinent formulations is the enforcement of Dirichlet boundary conditions \cite{zhu2019physics}. 
In our framework, these could be enforced a priori in $\by$, although depending on the representation of the solution, this is not always straightforward. A more elegant strategy would be to introduce the BCs at certain boundary points as actual observables through an additional likelihood (e.g. a Gaussian) which would ensure that the posterior on $\by$ satisfies them (up to some prescribed precision). The disadvantage of both of these procedures is that the approximate posterior, either $q_{\bpsi}(\by|\bx)$ or $q_{\bpsi}(\by_c|\bx)$ will also (approximately at least) enforce these Dirichlet BCs. Hence it would be useless in predictive settings where the PDE is the same but the Dirichlet BCs are different. 
In our formulation, Dirichlet BCs are enforced at the level of the implicit solver (i.e. on $\byy$) and are transferred to $\by$ through the appropriately fine-tuned linear map $\by=\bs{A}~\byy$ (see \ref{appendix:ytoYTransform}). In the multiscale setting where $\by_c$ inherits these Dirichlet BCs as described above, we set the corresponding  $\by_f$ (or $\by_f'$ in \refeq{eq:yf}) equal to $0$ a priori. As we demonstrate in the numerical illustrations, this enables us to produce extremely accurate predictions under BCs {\em not encountered} during training.

\item The hyperparameter $\lambda$ appearing in the ELBO and which controls the tightness of the tube that is implicitly defined in the joint $(\bx,\by)$  space around the $0$ residual manifold, offers a natural way to temper computations. One can envision starting with a small $\lambda$, which would correspond to an easier posterior to approximate as the residuals could be far from $0$ (in the limit that $\lambda=0$, the posterior coincides with the prior). Subsequently, this could be, potentially adaptively, increased until the target value is reached in order to attain posteriors (and approximations thereof) that are more tightly concentrated around the solution manifold. We defer investigations in this direction to future work.

\item Adaptive learning strategies could also be employed in terms of the samples $\{\bx_k\}_{k=1}^K$ employed during training. This is especially important in the multiscale setting as the dimension of the parameters $\{\by_{f,k}'\}_{k=1}^K$ that need to be fine-tuned is proportional to $K$. We defer investigations along this direction to future work.
\ei 

\begin{algorithm}[!t]
A set of random, fixed to the number inputs $\bx_k = \{\bx_1, \bx_2, ..., \bx_K \}$ is given:
\begin{algorithmic}[1]
    \State Select $\lambda$, $N$, $M$; Initialize $\bpsi\gets \boldsymbol{\psi_0}$;
    \State Construct a set of $N$ weighting functions \Comment{see subsection \ref{sub:probInference}}
    \State $\ell \gets 0$ \Comment{Initialize iteration counter}
    \While{$\mathcal{F}_{\bpsi}$ not converged}
        \State Draw $M$ weight functions at random \Comment{see \refeq{eq:randres}}
        \State Draw a subset of the K atoms $\bx_k$: $\bx_{k_s} \sim p(\bx), ~ k_s<K$
        \Comment{see Equation \eqref{eq:pemp}}
        \State  Generate $\bs{y}_{c,k_s} \gets \bs{A} \left( \byy\left(\bxx_{\bpsi_x}(\bx_{k_s}) +\bs{L}_Y \bs{\varepsilon}_1 +\sigma_Y \bs{\varepsilon}_2 \right) \right), \quad \bs{\varepsilon_1} \sim \mathcal{N}\left( \bs{0}, \bs{I}_{d_{\byy}'} \right), \quad \bs{\varepsilon_2} \sim \mathcal{N}\left( \bs{0}, \bs{I}_{d_{\byy}} \right)$ 
        \State {$\by \leftarrow \by_{c,k_s} + \bs{A}_{\perp} \by_{f,k_s}'$} 
        \State Approximate $\mathcal{F}_{\bpsi}$ \Comment{\textbf{Monte Carlo} Approximation according to Equation \eqref{eq:elboRandomResidualMultiscale}}
        \State Estimate $\nabla_{\bpsi} \mathcal{F}_{\bpsi}$ \Comment{\textbf{Backpropagation} for estimating $\nabla_{\bpsi} \mathcal{F}_{\bpsi}$}
        \State $\bpsi_{\ell+1} \gets \bpsi_{\ell} + \boldsymbol{\rho}^{(\ell)} \odot \nabla_{\bpsi} \mathcal{F}_{\bpsi}$ \Comment{\textbf{SGA} using \textbf{ADAM} for determining the learning rate $\bs{\rho}^{(\ell)}$}
        \State $\ell \gets \ell + 1$
    \EndWhile
\end{algorithmic}
\caption{mPANIS Training Algorithm.}
\label{alg:RandomResidualMulti}
\end{algorithm}

\subsection{mPANIS - Posterior Predictive Estimates}
\label{sec:mpanispred}
Upon convergence of the aforementioned SVI scheme and the identification of the optimal $\bpsi$ parameters, the approximate posterior $q_{\bpsi}(\by_c|\bx)$ in \refeq{eq:qyc}, i.e. the one accounting for the coarse-scale features of the full solution, can be used for posterior predictive estimates. For each test $\bx$, samples from $q_{\bpsi}(\by_c|\bx)$ can readily be drawn as in \refeq{eq:ycreparam}.
We note that for each such sample, the solution of the CG model (i.e. the determination of $\byy$ for the corresponding $\bxx_{\bpsi_x}(\bx)$) would be required.

In order to comparatively assess the performance of the proposed framework in the context of multiscale problems,  we consider a validation dataset $\mathcal{D}_v = \{ \mathbf{x}^j, \mathbf{u}^j \}_{j=1}^{N_{v}}$ consisting of $N_{v}$ inputs and discrete solution pairs, which were obtained by using a conventional FEM solver on a fine mesh.  From these we extract the coarse-scale part $\mathbf{u}_c^j=\sum_i^{d_{\by}} y_{c, i}^j\bs{\eta}_i$ of each $\mathbf{u}^j$ by projecting on the subspace spanned by the columns of the $\bs{A}$ matrix (i.e. $\by_c^j=\left(\bs{A}^T \bs{A}\right)^{-1} \bs{A}^T \by^j$), where $\bs{\eta}_i$ are the discretized equivalent of the shape functions described in Equation \eqref{eq:trialSolRepresentation}.

\noindent However, this is sufficient to approximate the solution well enough since proper uncertainty bounds will quantify the error. The detailed process for making predictions in the multiscale setup is shown in Algorithm \ref{alg:MakingPredictionsMulti}. Other than neglecting the influence of the fine-scale part, the procedure and the validation metrics are  identical to the ones described in Section \ref{sec:panispred}. 

\begin{algorithm}[!t]
    
\begin{algorithmic}[1]
    \State Compute  {$u_{c, \mu}(\bs{s}) \leftarrow \sum_{i=1}^{N_{\eta}} \left(\bs{A} \left( \byy\left(\bxx_{\bpsi_x}(\bx)\right)\right)\right)_i \eta_i(\bs{s})$} \Comment{Posterior Mean of Solution Field };
    \State Compute  {$u_{c, \sigma}(\bs{s}) \leftarrow \sqrt{ \sum_{i,j=1}^{N_{\eta}}  \eta_i(\bs{s}) \bs{\Sigma}_{\bpsi, ij} \eta_j(\bs{s}) }$} \Comment{Posterior St. Deviation of Solution Field}
  
    \State $u_{c, U}(\bs{s}) \leftarrow u_{c, \mu}(\bs{s}) + 2 u_{c, \sigma}(\bs{s})$ \Comment{Upper Uncertainty Bound Field}
    \State $u_{c, L}(\bs{s}) \leftarrow u_{c, \mu}(\bs{s}) - 2 u_{c, \sigma}(\bs{s})$ \Comment{Lower Uncertainty Bound Field}
\end{algorithmic}

\caption{mPANIS Posterior Predictive Estimates for coarse-scale solution  (Equation \eqref{eq:trialSolRepresentation}) for a test input $\bx$}
\label{alg:MakingPredictionsMulti}
\end{algorithm}

\section{Numerical Illustrations}
\label{sec:numerical}
This section contains several numerical illustrations of the performance of the proposed (m)PANIS framework on linear and nonlinear PDEs. It is compared with the  state-of-the-art method of Physics Informed Fourier Neural Operators (PINO) \cite{li2021physics}. In parallel, components of the model generically described in the previous sections are specified, and additional implementation details are provided. Our objectives are to illustrate:
\bi
\item the comparative performance under varying dimensions of the parametric input (\textbf{predictive accuracy}).

\item the very low number of training parameters of the proposed architectures in relation to competitors (\textbf{lightweight}).

\item the ability of the proposed framework to generalize i.e. to provide accurate predictions in out-of-distribution settings as well as under different boundary conditions (\textbf{generalization}).

\item the ability to capture the  low-frequency components of the solution in multiscale PDEs (\textbf{multiscale}).
\ei

\begin{figure}[!t]
    \centering
    \hfill
    \begin{minipage}[b]{0.4\textwidth}
        \centering
        \includegraphics[width=\textwidth]{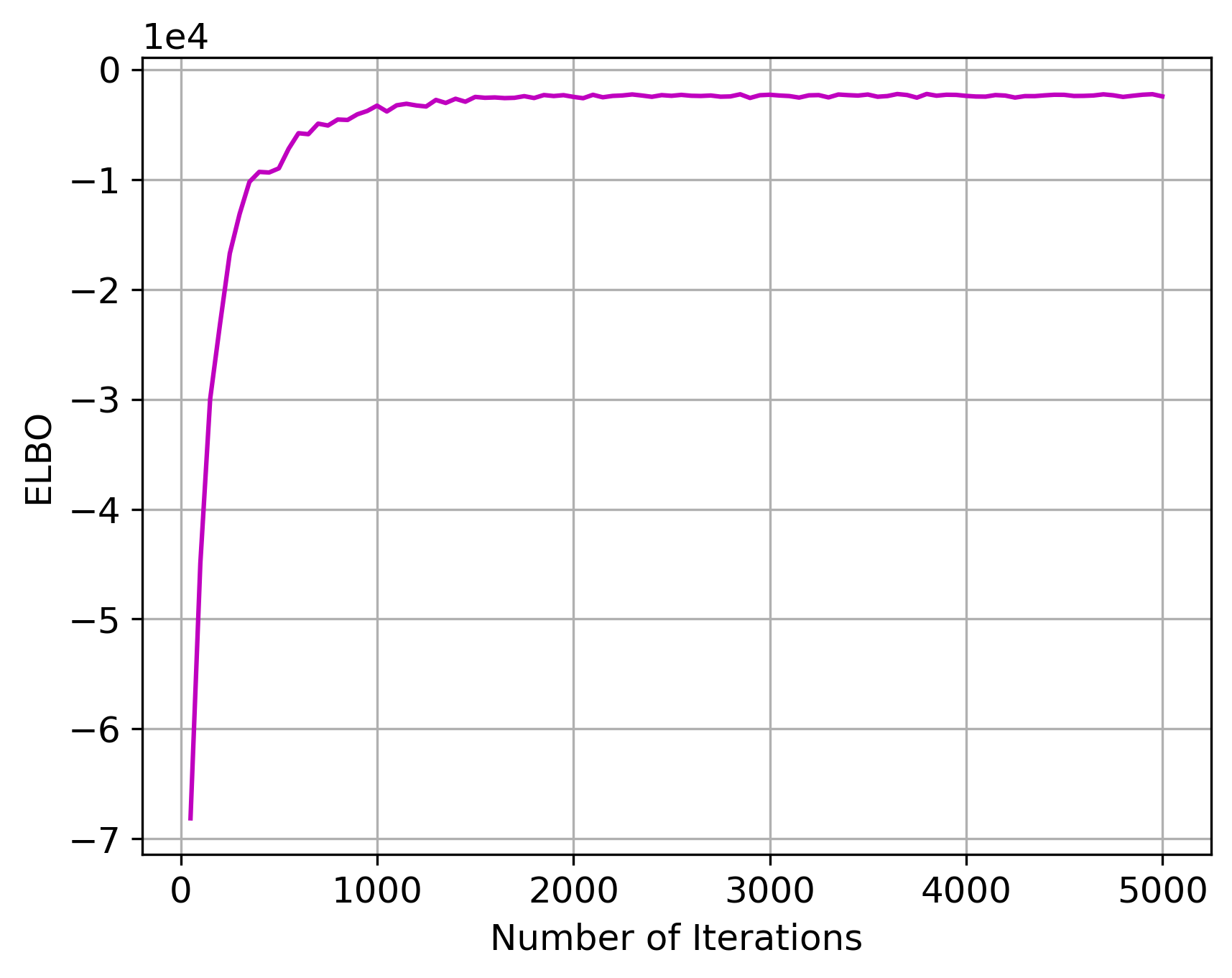}
    \end{minipage}
    \hfill
    \begin{minipage}[b]{0.29\textwidth}
        \centering
        \includegraphics[width=\textwidth]{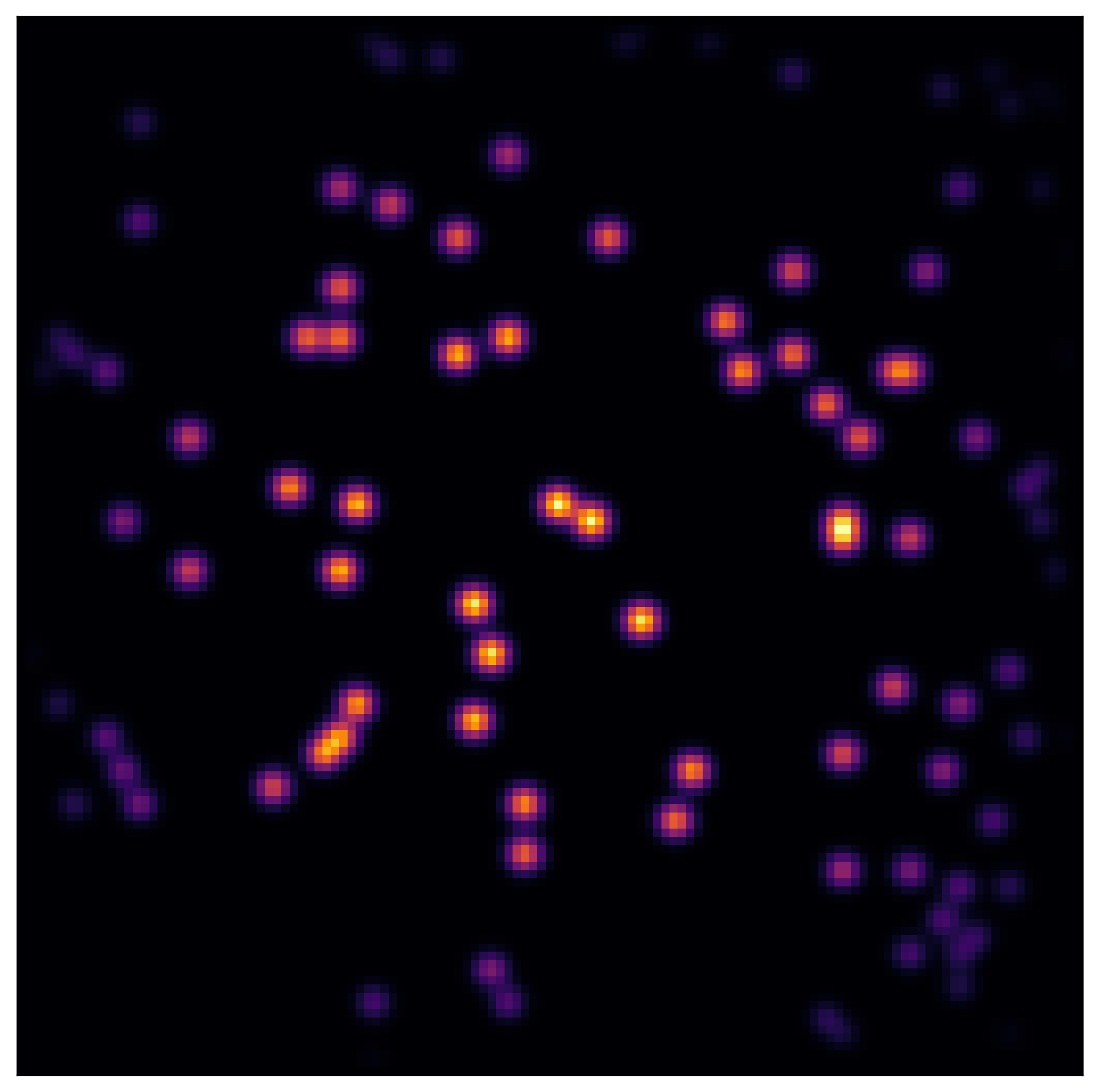}
    \end{minipage}
    \hfill
    \begin{minipage}[b]{0.29\textwidth}
        \centering
        \includegraphics[width=\textwidth]{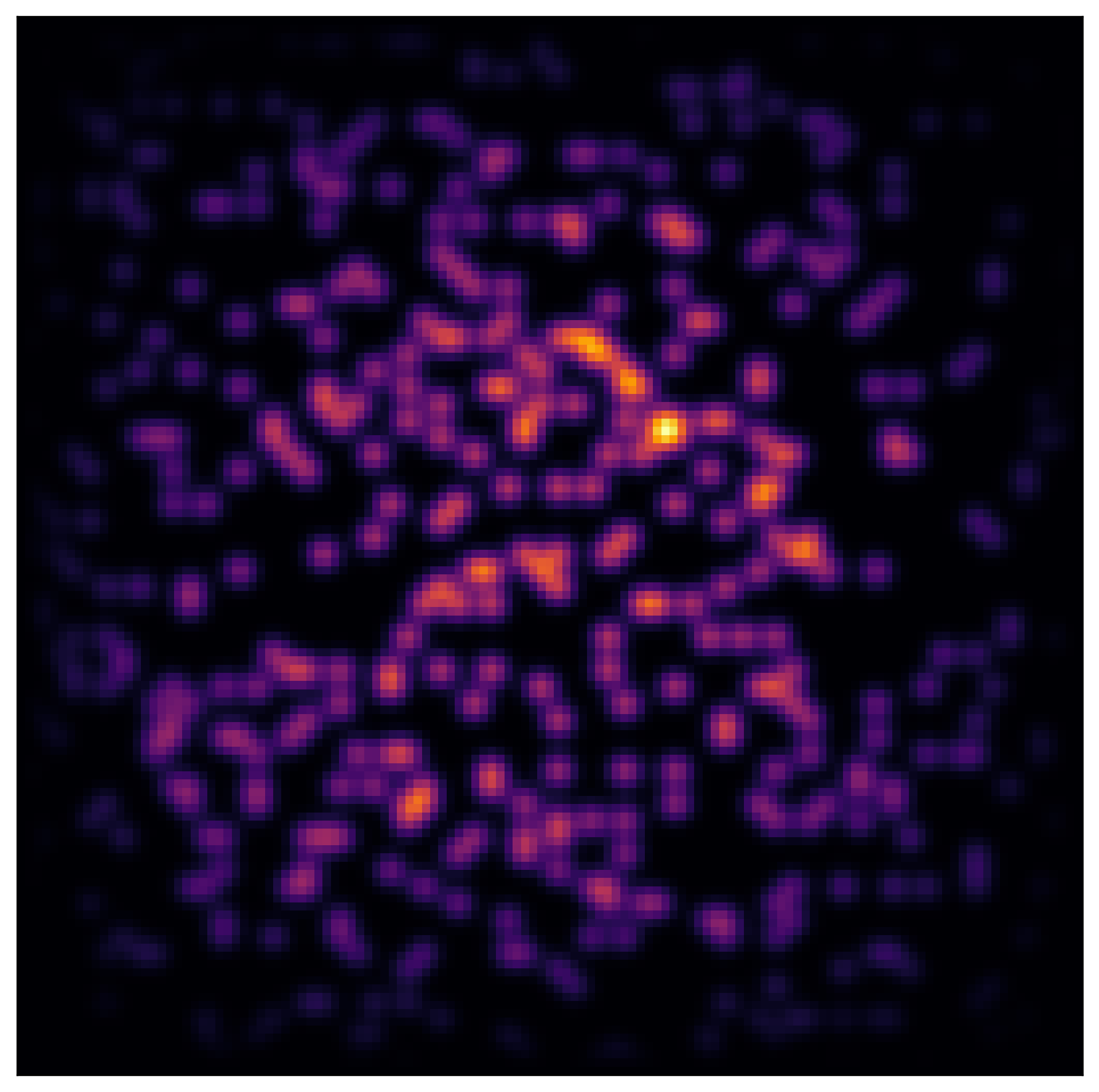}
    \end{minipage}
    \caption{Convergence of the ELBO $\mathcal{F}$ for PANIS, when trained on microstructures with $VF=50\%$ and BCs $u_0=0$ (left). Illustration of 100 and 500 randomly selected RBF-based weight functions $w_j$ according to algorithm \ref{alg:RandomResidual} (middle-right). Each one of these $w_j$'s  corresponds to a single weighted residual.}
    \label{fig:randomlySelectedRbfs}
\end{figure}

The general implementation of PANIS and the dedicated coarse-grained solver mentioned in subsection \ref{sec:ApproximatingDensity} is based on PyTorch \cite{paszke2019pytorch}. Training was conducted on an H100 Nvidia GPU and takes $\sim 10 ~min$ for the ELBO to converge (see Figure \ref{fig:randomlySelectedRbfs}) in the case of PANIS and $\sim 2195 ~min$ in the case of PINO. The validation dataset in all the subsequent illustrations consists of $N_v=100$ input-solution pairs.  Reference solutions were obtained using Fenics \cite{fenics2015} and a uniform, finite element mesh with $57600$ nodal points. 

A github repository containing the associated code and illustrative examples will become available upon publication at \href{https://github.com/pkmtum}{https://github.com/pkmtum/PANIS}.

\subsection{High Dimensional Darcy Problem} \label{sub:2D_Darcy}
We begin with the Darcy flow equation: 

\begin{equation}
\begin{array}{lll}
 \nabla \cdot \left( - c(\bs{s};~ \bx) \nabla  u(\bs{s})\right) = f, ~~~ \mathbf{s} \in \Omega = \left[0, ~ 1 \right] \times\left[0, ~ 1 \right] \\
u(\bs{s}) = u_0, ~~~ \mathbf{s} \in \Gamma_D
\end{array}
\label{eq:2D_Darcy}
\end{equation}
as this PDE has been considered by the majority of state-of-the-art methods \cite{li2020fourier} \cite{li2021physics} \cite{lu2022comprehensive} and the respective code is generally available for producing the following comparative results. In the following, we assume that $f=100$, $u_0=0$, and the input $\bx$ parametrizes the permeability field $c(\bs{s};~ \bx)$ as described in the sequel. 

\textbf{Input permeability $c(\bs{s};~\bx)$:}
 The permeability fields are generated as cut-offs from an underlying Gaussian field. In particular we consider a zero-mean Gaussian field $G(\bs{s}; ~\bx)=\sum_{i=1}^{d_x} \sqrt{\lambda_i} x_i v_i(\bs{s})$, where $x_i \sim \mathcal{N}(0,1)$ determine the input vector $\bx$, $\lambda_i$ are the eigenvalues and $v_i(\bs{s})$ the eigenfunctions of the covariance kernel  $k(\mathbf{s}, \mathbf{s}^{\prime}) = exp\left(-\frac{||\mathbf{s}-\mathbf{s}^{\prime}||^2_2}{l^2}\right)$ where  $l$ determines  the length scale \cite{zhu2019physics}. By controlling the numbers of terms $d_{\bx}$, we can control the dimension of the parametric input.
 We generate  $c(\bs{s}; \bx)$ corresponding to a binary medium as follows:

\be
c(s_i, \bx) = \begin{cases} 1, ~~\textrm{if }~G(s_i ; \bx) \ge t ~~~\textrm{(phase 1)} \\ \frac{1}{CR}, ~~\textrm{if }~G(s_i; \bx)< t  ~~~\textrm{(phase 2)}\end{cases}
\label{eq:condField}
\ee
where the cut-off threshold is set as $t=\Phi^{-1}(VF)$ where $VF$ is the prescribed volume fraction of phase 1 and $\Phi$ is the standard Gaussian CDF (e.g. for $VF=0.5$, $t=0$). 
The parameter $CR$ defines the contrast ratio in the properties of the phases, which has a significant impact on the medium's response as well as on the construction of pertinent surrogates. In general, the higher $CR$ is, the more higher-order statistics affect the response \cite{torquato2002random}. 
We employed a $CR=10$ in the subsequent illustrations, although much smaller values are generally selected \cite{lu2022comprehensive, bhattacharya2021model}. 

\begin{figure}[!t]
    \centering
    \includegraphics[width=0.95\linewidth,height=0.3\textheight]{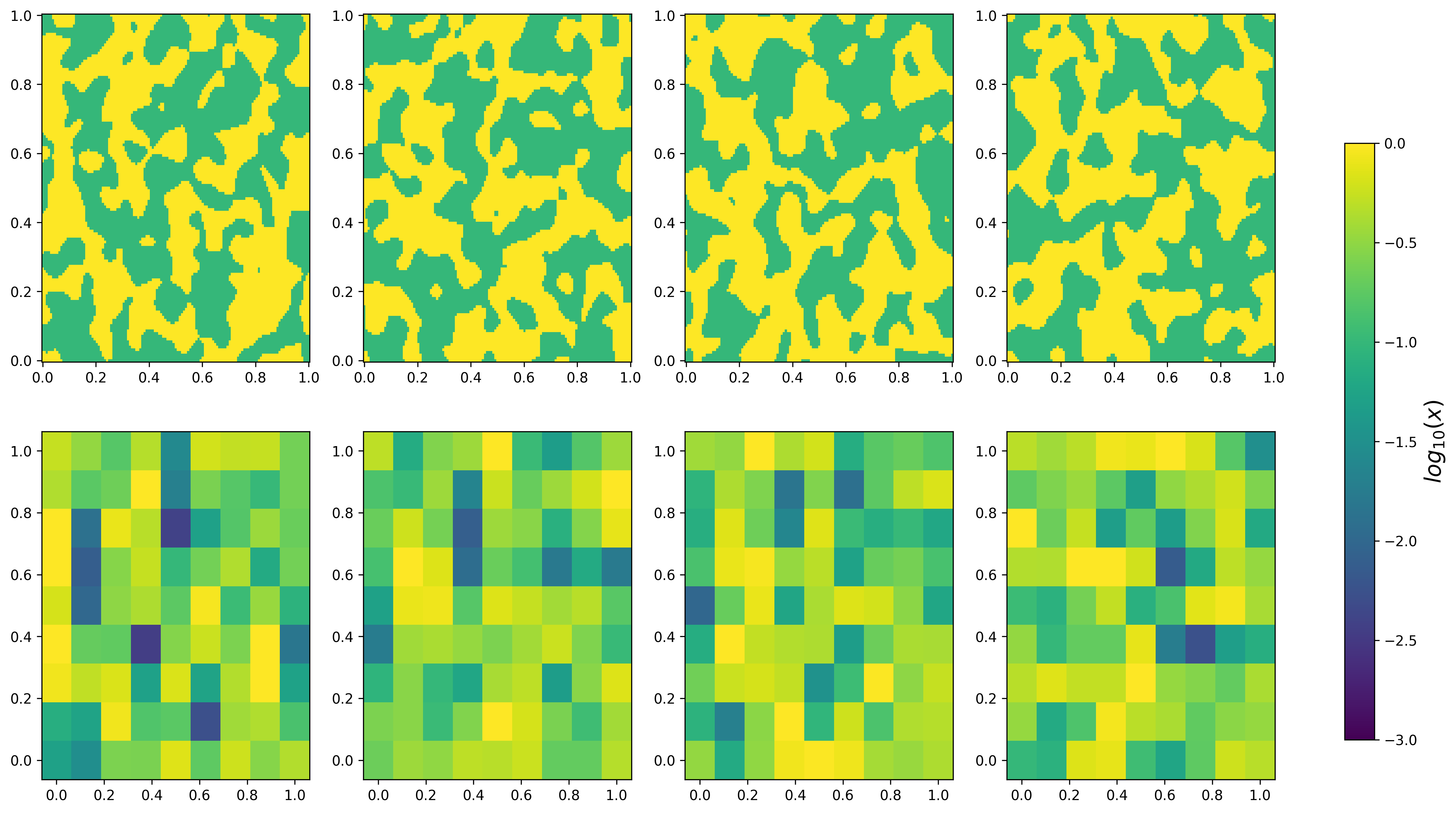}
    \caption{Indicative pairs of the full permeability field $c(\bs{s}; \bx)$ (generated for  $l=0.05$ as described in section \ref{{sub:2D_Darcy}}) and the learned coarse-grained input $\bxx$.}
    \label{fig:xXpairs}
\end{figure}

As mentioned in subsection \ref{sec:ApproximatingDensity}, the first layer of PANIS architecture (see Equation \eqref{eq:formOfmeanPrediction}) involves the fine-to-coarse map denoted by $\bxx_{\bpsi_x}(\bx)$ that is realized as a shallow CNN (Appendix \ref{appendix:shallowCNN}). We note at this point that the input of the CNN is the conductivity field $c(\bs{s};~ \bx)$ (i.e. implicitly $\bx$) obtained from Equation \eqref{eq:condField} and the output are the effective properties $\bxx$. Some intuition about the role of $c(\bs{s};~ \bx)$ and $\bxx$ could be drawn from Figure \ref{fig:xXpairs}, which visualizes some indicative pairs of $c(\bs{s};~\bx)$ and $\bxx$. For each full conductivity field $c(\bs{s};~ \bx)$, the model learns a coarse-grained equivalent $\bxx$, which, when combined with the rest of the model components, scores best in terms of the weighted residuals. In general, areas with high/low $c(\bs{s}; ~\bx)$ yield areas with high/low $\bxx$-values respectively, although the ranges can differ.

\textbf{Trial solutions and weighting functions:} The solution $u$ is represented as in \refeq{eq:trialSolRepresentation} with:
\be
\eta_i(\bs{s})=exp\left( \frac{|| \bs{s} - \bs{s_{i,0}}||^2_2}{\Delta l^2} \right)
\ee

In total, $d_{\by}=4096$ (i.e. $dim(\by)=4096$) functions were used. 
The center-points  $\bs{s_{i,0}}$ were located on a regular $64\times 64$ grid over the problem domain. As mentioned earlier, BCs are enforced indirectly through the implicit solver (i.e. through $\byy$ in \refeq{eq:formOfmeanPrediction}).

For the population of $N$ weighting functions $w_j(\bs{s})$ that define the corresponding weighted residuals in  \refeq{eq:virtuallike}, the same functions $\eta_j(\bs{s})$ above were employed with the following differences for PANIS and mPANIS:

\begin{itemize}
    \item For PANIS, $N=289$ and the center-points $\bs{s_{i,0}}$ were located on a regular grid $17 \times 17$.
    \item For mPANIS, $N=4096$ and the center-points $\bs{s_{i,0}}$ were located on a regular grid $64 \times 64$.
\end{itemize}
These were multiplied with the function $\tau(s)=s_1(1-s_1)~s_2(1-s_2)$, which is merely used to enforce that they are zero on the Dirichlet boundary \footnote{Other such functions could be readily employed.}, i.e.:
\begin{equation}
   w_j(\bs{s})=\tau(\bs{s})~\eta_j(\bs{s}).
   \label{eq:weightFuncs}
\end{equation}

We note that in the framework proposed, the selection of the weighting function space is detached from the trial solutions and the discretization of $u$. Naturally, and as each $w_j$ furnishes different amounts of information about the solution, an intelligent selection could significantly impact the efficiency of the method, but it is not investigated in this work. Numerical integrations over the problem domain for the evaluation of the associated weighted residuals (\refeq{eq:weak1}) are performed using the trapezoidal rule on a regular $128 \times 128$ grid.

\textbf{Coarse-Grained Implicit Solver:}
The coarse-grained implicit solver, which lies at the center of the proposed architecture, is obtained by employing a regular, linear, finite element discretization grid of $16 \times 16$ triangular pairs for PANIS and $8 \times 8$ for mPANIS. Each of the resulting $512$ (or $128$) triangular elements is assumed to have a constant permeability, which is summarily represented by the vector $\bxx$ (\refeq{eq:formOfmeanPrediction}). The solution of the corresponding discretized equations gives rise to the nodal values which are represented by the vector $\byy$ (\refeq{eq:formOfmeanPrediction}).

\begin{table}[!t]
  \centering
  \begin{tabular}{|c|c|c|c|}
    \hline
    Variable & PANIS & mPANIS & PINO \\
    \hline
    $\mathbf{x}$ & $1024$ & $1024$ & --- \\
    \hline
    $\mathbf{y}$ & $4096$ & $4096$ & --- \\
    \hline
    $\mathbf{X}$ &  $289$ & $81$ & --- \\
    \hline
    $\mathbf{Y}$ & $289$ & $81$ & --- \\
    \hline
(model parameters)    $\bpsi$ & $7956$ & $415112$ & $13140737$ \\
    \hline
    $\bpsi_x$ & $5065$ & $ 12801$ & --- \\
    \hline
    Wall Clock Time (min) & $\approx 10$ & $\approx 807$ & $\approx 2195$ \\
    \hline
    Resolution of $c(\bs{s};~ \bx)$ & $128$ & $128$ & $128$ \\
    \hline
    Resolution of $u(\bs{s})$ & $128$ & $128$ & $128$ \\
    \hline
  \end{tabular}
  \caption{Most important dimensions of PANIS, mPANIS and PINO.}
  \label{tab:Dims}
\end{table}

A summary of the most important dimensions of the models considered is contained in Table \ref{tab:Dims}.

\subsubsection{Comparison for Different Volume Fractions}
\label{sub:diffVFs}
In the following illustrations the predictive accuracy of PANIS will be demonstrated and compared against PINO for various volume fractions $VF$. Both models were trained for input fields $c(\bs{s};~\bx)$ corresponding to $VF=50~ \%$. Subsequently, their predictive performance was tested for $c(\bs{s}; \bx)$ corresponding to a range of different VFs. The convergence of the ELBO during training of PANIS is depicted in Figure \ref{fig:randomlySelectedRbfs}. Lastly, for all the subsequent illustrations presented in subsections \ref{sub:2D_Darcy} and \ref{sub:nonLinear} the hyperparameters of Algorithm \ref{alg:RandomResidual} are set $\lambda=10^{4}$, $N=289$ and $M=100$ ($M=200$ for subsection \ref{sub:nonLinear}).

\begin{figure}[!h]
    \centering
    \includegraphics[width=\linewidth]{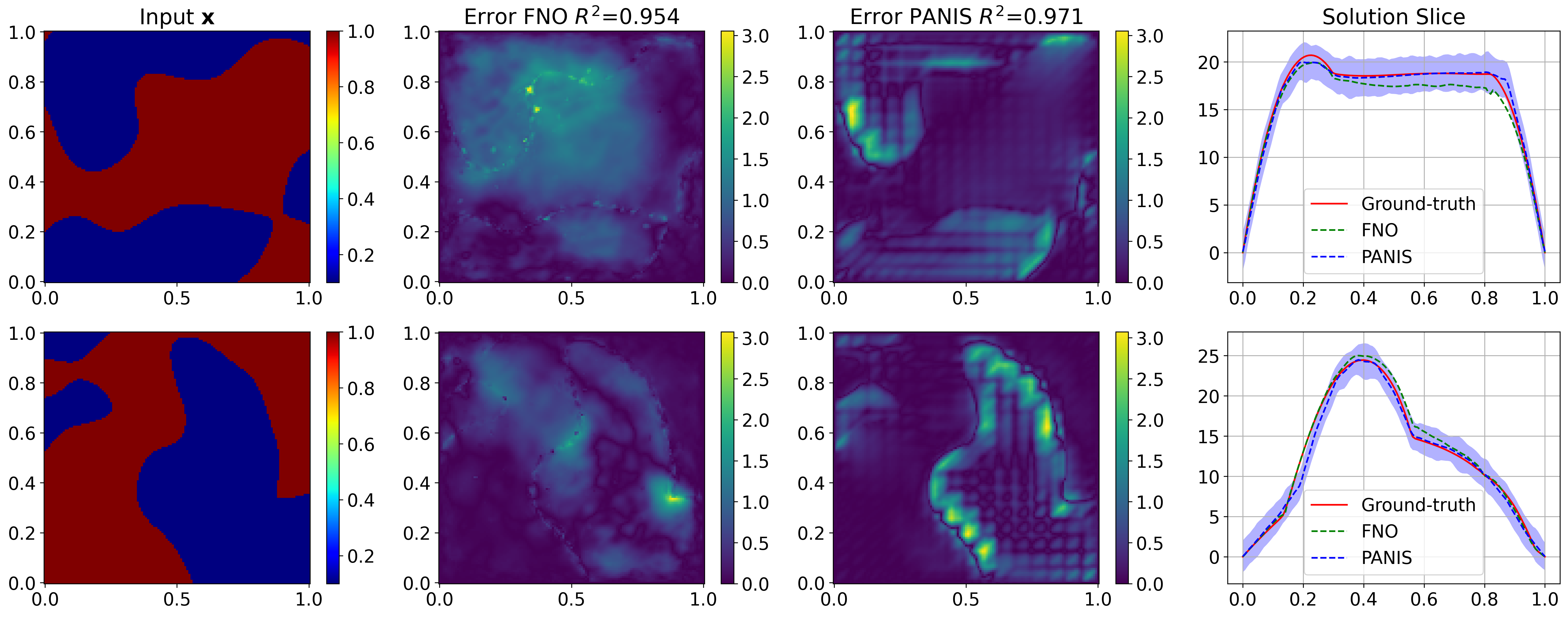}
    \caption{Predictive accuracy of FNO and PANIS when trained and tested on microstructures with $VF=50 ~\%$. The right column shows a one-dimensional slice of the solution along the vertical line from (0.5,0) to (0.5, 1.0). The shaded blue area corresponds to $\pm 2$ posterior standard deviations computed as described in Algorithm \ref{alg:MakingPredictions}. }
    \label{fig:CompFNO}
\end{figure}

We note that PINO gives better predictions when it is used in a hybrid mode, i.e. when both data-driven and physics-informed terms are used in the loss function \cite{li2021physics}.
In the remainder of the paper, we focus purely on physics-informed versions for a fair comparison. Nevertheless, we have conducted one comparison between PANIS and the purely data-based  FNO  (i.e. the data-based version of PINO) in Figure \ref{fig:CompFNO} to demonstrate the competitiveness of PANIS even against well-established data-driven methods such as FNO. 

\begin{figure}[!h]
    \centering
    \subfigure[Predictive accuracy of PANIS and PINO when trained and tested on microstructures with $VF=50 ~\%$.]{
        \includegraphics[width=0.9\linewidth]{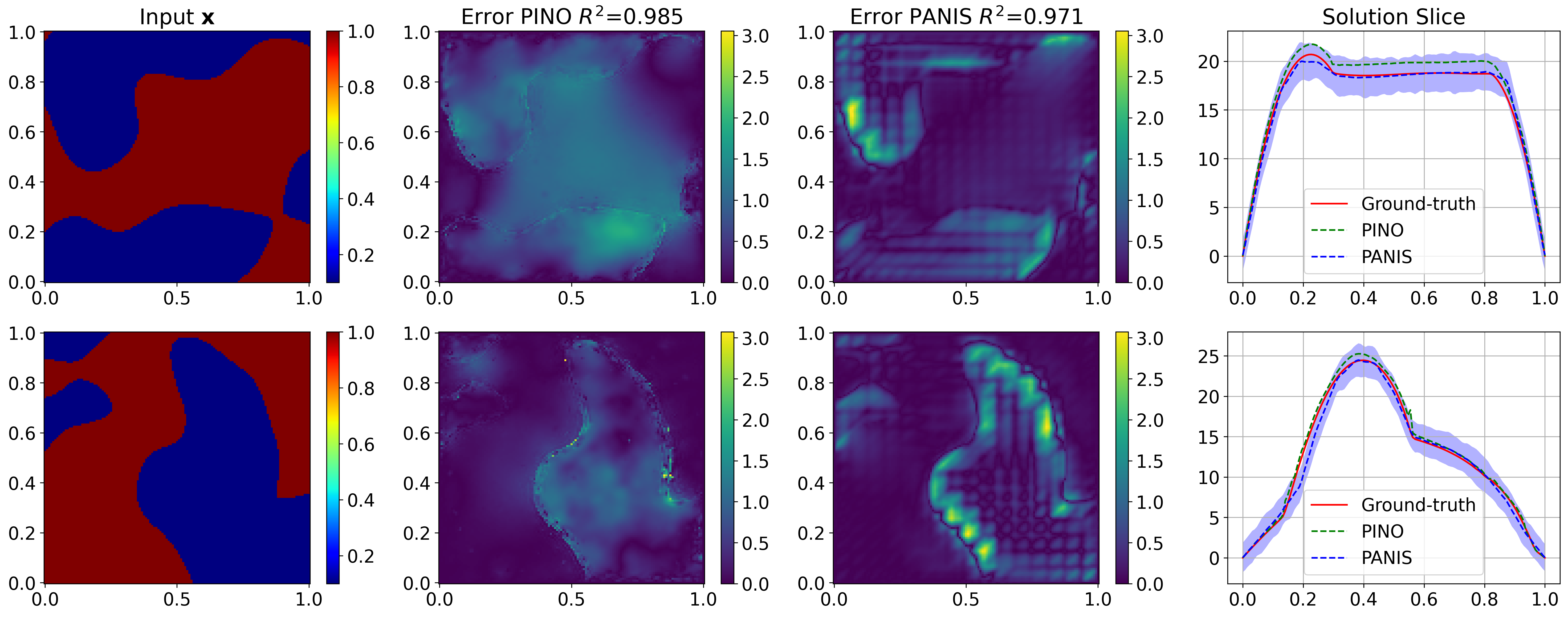} 
    }
    
    \vspace{1cm} 
    \subfigure[\textbf{Out of distribution Prediction:} Predictive accuracy of PANIS and PINO when trained on microstructures with $VF=50~\%$ and tested on $VF=10 ~\%$.]{
        \includegraphics[width=0.9\linewidth]{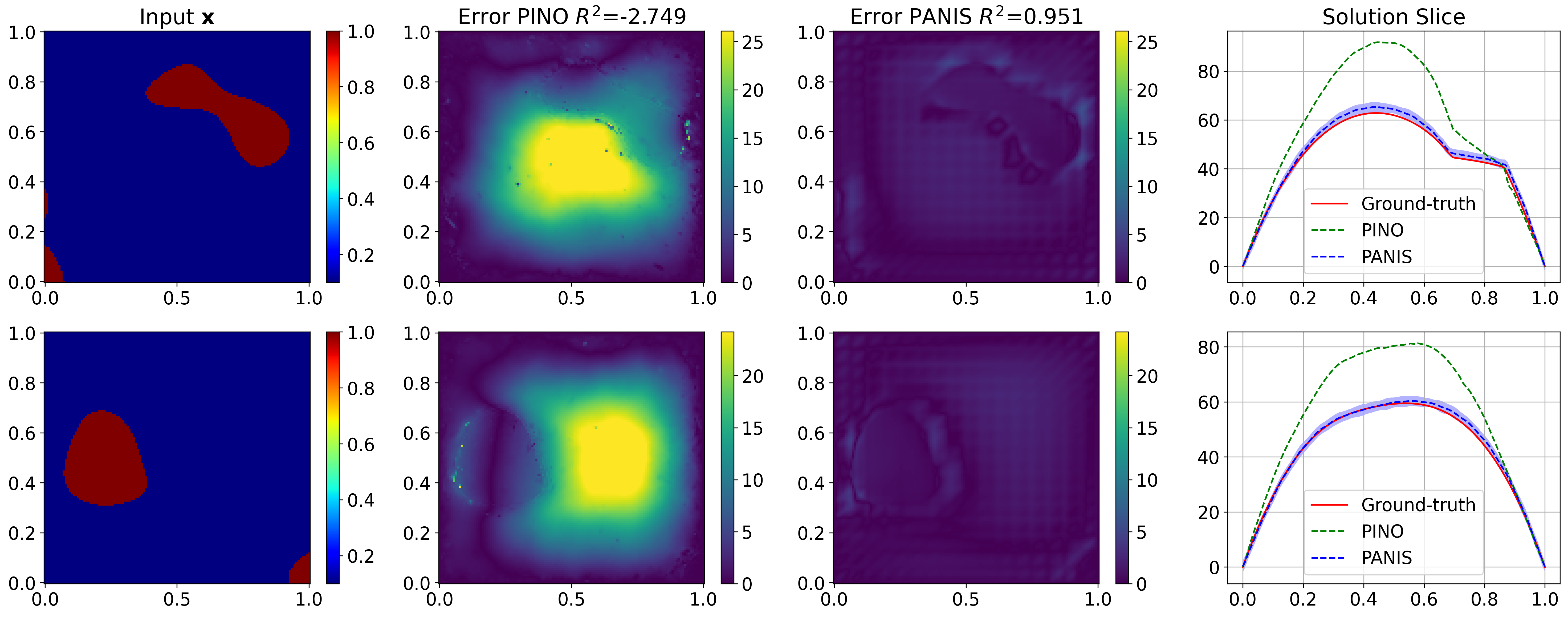}
    }
    
    \caption{Comparison  between PANIS and PINO.}
    \label{fig:VFcompNormal}
\end{figure}

The performance of the two models (PANIS and PINO) is similar for in-distribution predictions, as can be seen in Figure \ref{fig:VFcompNormal}. The mean prediction in both cases is close to the ground truth, as seen from the coefficient of determination $R^2$. However, PANIS provides uncertainty bounds for the predicted solution, in contrast to PINO, which yields only point estimates for the PDE solution.
When both models are inquired in out-of-distribution conditions (i.e. different volume fraction $VF$), PANIS fully retains its predictive accuracy  in contrast to PINO which struggles as seen in Table \ref{tab:DiffVFs}. The latter  depicts the $R^2$ score under various $VF$s. Due to the physics-aware, implicit solver which lies at the center of its architecture,   PANIS  is capable of retaining the  predictive accuracy for every  $VF$ examined, as opposed to PINO, whose accuracy decreases significantly as the volume fraction $VF$ deviates notably from the one used for training, i.e. $VF=50\%$.

\subsubsection{Comparison for Different Dimension of the parametric input $\bx$}
In addition, we examined the comparative performance of the aforementioned  models for different dimensions $d_{\bx}$ of the parametric input $\bx$, which affects the permeability field as explained in Section \ref{sub:2D_Darcy}. As it can be seen from Table \ref{tab:compInput}, although the predictive accuracy of both models decreases as the parametric input increases, it remains very high even when the parametric input is $1024$.  We emphasize, however, that the number of training parameters $\bpsi$ for PANIS is $O(10^4$) in contrast to PINO, which employs approximately $O(10^7)$ (Table \ref{tab:Dims}). The utilization of the coarse-grained implicit solver enables the reduction of complexity and leads to comparably good results with  approximately \textbf{three orders of magnitude} fewer parameters.

\begin{table}[!t]
    \centering
  \begin{tabular}{|c|c|c|}
    \hline
    Volume Fraction (VF) & $R^2$ (PINO) & $R^2$ (PANIS) \\
    \hline
    $10 ~\%$ & $-2.749$ & $0.951$ \\
    \hline
    $20 ~\%$ & $0.510$ & $0.962$ \\
    \hline
    $30 ~\%$ & $0.911$ & $0.969$ \\
    \hline
    $40 ~\%$ & $0.967$ & $0.970$ \\
    \hline
    $50 ~\%$ (training) & $0.985$ & $0.971$ \\
    \hline
    $60 ~\%$ & $0.988$ & $0.964$ \\
    \hline
    $70 ~\%$ & $0.983$ & $0.963$ \\
    \hline
    $80 ~\%$ & $0.968$ & $0.953$ \\
    \hline
    $90 ~\%$ & $0.911$ & $0.928$ \\
    \hline
  \end{tabular}
  \caption{Predictive accuracy in terms of $R^2$ (the closer to $1$ it is, the better) for PANIS and PINO when tested on various $VF$ values while trained only on $VF=50~\%$.}
  \label{tab:DiffVFs}
\end{table}

\begin{table}[!t]
\centering
  \centering
  \begin{tabular}{|c|c|c|}
    \hline
    $d_x$ (dimension of $\bx$) & $R^2$ (PINO) & $R^2$ (PANIS) \\
    \hline
    64 & 0.988 & 0.982 \\
    \hline
    256 & 0.987 & 0.979 \\
    \hline
    1024 & 0.985 & 0.971 \\
    \hline
  \end{tabular}
  \caption{Coefficient of determination $R^2$ for PINO and PANIS for various parametric inputs $\bx$, when trained on $VF=50 ~\%$, $u_0=0$.}
  \label{tab:compInput}
\end{table}

\subsubsection{Comparison for Different Boundary Conditions}
\label{sub:diffBCs}
In the following illustrations, we focus on a challenging test for the generalization capability of any data-driven scheme. In particular, we report the  accuracy of the trained models when called upon to make predictions under different  boundary conditions from the ones used to train the models. For this purpose, we train both models with the same conditions as in \refeq{eq:2D_Darcy} i.e. $u_0=0$, and we test them for 2 very different Dirichlet boundary conditions. 
The first corresponds merely to an offset the BCs i.e. for $u_0=10$. The comparison is conducted in terms of relative $L_2$ error $\epsilon$ (\refeq{eq:relError}) as shown in Table \ref{tab:diffBCs}. We can observe that PINO fails to predict that the solution simply needs to be shifted by $10$ which leads to a high $\epsilon$-value. In contrast, PANIS retains intact its predictive accuracy.

The second  boundary condition considered deviates much more from $u_0=0$ used for training, and in particular, we employed:

\begin{equation}
u_0(s_1, s_2) =
\begin{cases}
    10 + 5 \sin\left(\frac{\pi}{2} s_2\right), & \text{if } s_1 = 0, \quad 0 \leq s_2 \leq 1, \\
    10 + 5 \sin\left(\frac{\pi}{2}(s_1+1)\right), & \text{if } s_2=1, \quad 0 \leq s_1 \leq 1, \\
    10 - 5 \sin\left(\frac{\pi}{2}(s_2 + 1)\right), & \text{if } s_1 = 1, \quad 0 \leq s_2 \leq 1, \\
    10 - 5 \sin\left(\frac{\pi}{2}(s_1)\right), & \text{if } s_2 = 0, \quad 0 \leq s_1 \leq 1.
\end{cases}
\label{eq:sinBCs}
\end{equation}

\begin{figure}[!h]
    \centering
    \includegraphics[width=0.9\linewidth]{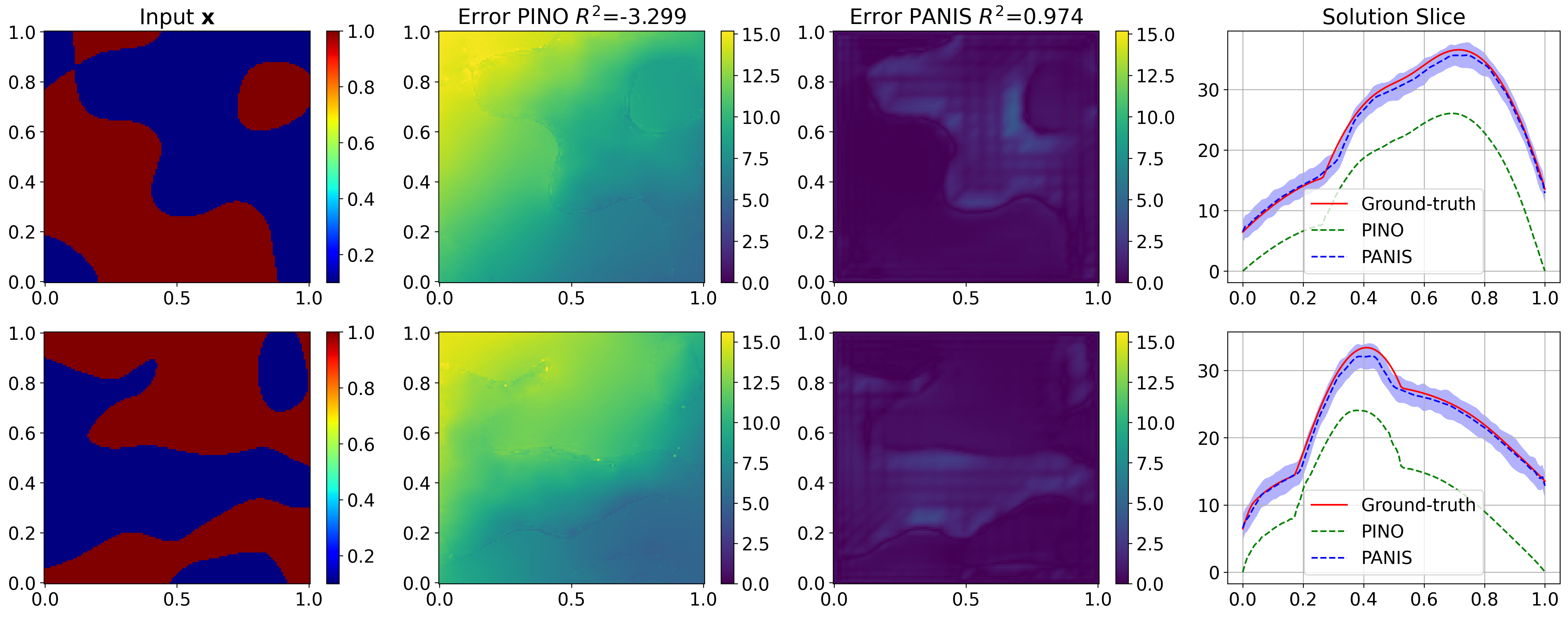}
    \caption{Comparison between PANIS and PINO when trained on $VF= 50~\%$, $u_0=0$ and tested on $VF=~50 \%$, $u_0$ as described in Equation \eqref{eq:sinBCs}.}
    \label{fig:diffBCs}
\end{figure}

Indicative  predictions provided by both models  for this non-trivial case are presented in Figure \ref{fig:diffBCs}. In Table \ref{tab:diffBCs}, we observe that as in the previous case, and due to the embedded physics,  the performance of PANIS is not  affected by the change of boundary conditions, whereas the predictions of PINO deteriorate drastically.

\begin{table}[!h]
  \centering
  \begin{tabular}{|c|c|c|}
    \hline
    Boundary Condition Type & $\epsilon$ (PINO) & $\epsilon$ (PANIS) \\
    \hline
    $u_0=0$  (training) & $0.0381$ & $0.0589$ \\
    \hline
    $u_0=10$  & $0.4581$ & $0.0368$ \\
    \hline
    $u_0$ as in Equation \ref{eq:sinBCs}  & $0.4763$ & $0.0347$ \\
    \hline
  \end{tabular}
  \caption{Comparison of relative $L_2$ error $\epsilon$ (the smaller, the better) when tested under different boundary conditions.}
  \label{tab:diffBCs}
\end{table}

\subsection{Non-Linear Poisson Equation}
\label{sub:nonLinear}
In this subsection, we assess the performance of the proposed model on a non-linear PDE. We note that  no adaptations or changes are needed as the sole conduit of information is the weighted residuals. As a result, the training algorithm merely requires the computation of the residuals and  their derivatives with respect to the model parameters (with automatic differentiation). 

We consider the same conservation law, which in terms of the flux vector $\bs{q}$, can be written as:
\begin{equation}
\begin{array}{lll}
\nabla \cdot  \bs{q}  = f, ~~~ \mathbf{s} \in \Omega = \left[0, ~ 1 \right] \times\left[0, ~ 1 \right] \\
\end{array}
\label{eq:nonLinearPDE}
\end{equation}
which is complemented by a  {\em non-linear} constitutive law of the form:
\begin{equation}
\bs{q}(\bs{s}) = - c(\bs{s}; ~\bx) e^{\alpha (u(\bs{s}) - \bar{u})}~ \nabla u(\bs{s}).
\label{eq:nonlincl}
\end{equation}
and Dirchlet boundary condition $u=u_0, ~~\mathbf{s} \in \Gamma_D$ as before.  
 \noindent The  field $c(\bx, \bs{s})$ is defined  as described in subsection \ref{sub:2D_Darcy}, and $\alpha$, $\bar{u}$\footnote{The values $a=0.05$ and $\bar{u}=5$ were used in the subsequent numerical results.} are two additional scalar parameters (assumed to be independent of $\bs{s}$). It is noted that $\alpha$ (and secondarily $\bar{u}$) controls the degree of non-linearity. For $\alpha = 0$, Equation \ref{eq:nonLinearPDE} will take the same form as in Equation \ref{eq:2D_Darcy}, and as $\alpha$ increases, so does the nonlinearity in the governing PDE. 

The most significant adjustment pertains to the implicit solver that lies at the center of the architecture proposed. Apart from being lower-dimensional and more efficient to solve, one is at liberty to induce as much physical insight and complexity \cite{grigo_physics-aware_2019}. In this work, we adopt the same discretization explained before and a constitutive law of the same nonlinear form as in \refeq{eq:nonlincl}, where in each element of the coarse-grained model  instead of $c(\bs{s}; \bx)$ we have the learnable parameters denoted by $\bxx$ (i.e. $a$ and $\bar{u}$ have the same values). We note that one could adopt a different form for the constitutive law or make the corresponding $a,\bar{u}$ learnable as well.  
In order to solve the corresponding non-linear system of equations, i.e. to determine $\byy(\bxx)$,  we  employed Newton's iterative method. This was achieved in a fully vectorized way, such that PyTorch can backpropagate through the computational graph  for computation of the gradients  needed for estimating the derivatives of the ELBO.
In order to further expedite the computations, we initialized the model parameters to the ones found for the linear case, i.e. $a=0$ (section \ref{sub:2D_Darcy}) and subsequently increased the $a$ gradually per iteration until the final value of $a=0.05$ was reached. This provides a natural tempering mechanism that can expedite and stabilize convergence. In addition, and if this is of interest, it enables one to automatically obtain surrogates for intermediate values of the constitutive parameter $a$ which could be used for predictive purposes or sensitivity analysis.

\begin{figure}[!h]
    \centering
    \includegraphics[width=0.85\linewidth]{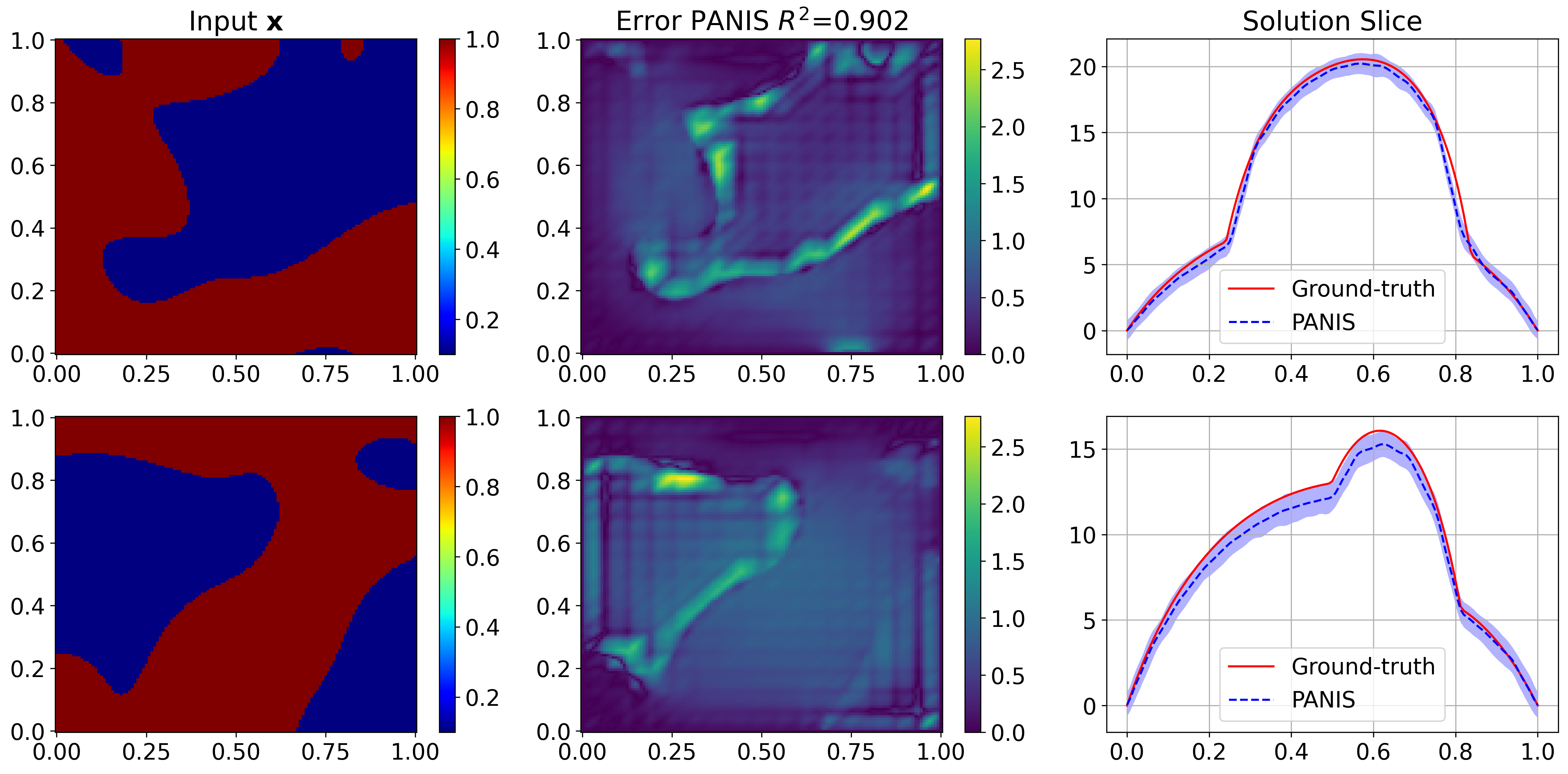}
    \caption{Predictions obtained with PANIS  for the non-linear PDE of  \refeq{eq:nonLinearPDE} and microstructures generated with  $l=0.25$, $VF=50~\%$ and $u_0=0$.}
    \label{fig:nLinear}
\end{figure}

In Figure \ref{fig:nLinear}, we show the predictive performance of PANIS for two indicative samples. 
The predictive accuracy remains high as previously, demonstrating that the proposed framework can also be used effectively in non-linear, parametric  PDEs. In out-of-distribution settings as those obtained by altering the   volume fraction or the boundary conditions (see sections \ref{sub:diffVFs} and \ref{sub:diffBCs}), PANIS retains intact its prediction accuracy as it can be seen in  Table \ref{tab:nonLinear}.

\begin{table}[t]
  \centering
  \begin{tabular}{|c|c|c|}
    \hline
    Volume Fraction (VF) & Boundary Condition Type & $\epsilon$ (PANIS) \\
    \hline
    $10 ~\%$ & $u_0=0$ everywhere & $0.0541$ \\
    \hline
    $20 ~\%$ & $u_0=0$ everywhere & $0.0714$ \\
    \hline
    $30 ~\%$ & $u_0=0$ everywhere & $0.0770$ \\
    \hline
    $40 ~\%$ & $u_0=0$ everywhere & $0.0854$ \\
    \hline
    $50 ~\%$ (training) & $u_0=0$ everywhere & $0.0904$ \\
    \hline
    $60 ~\%$ & $u_0=0$ everywhere & $0.0966$ \\
    \hline
    $70 ~\%$ & $u_0=0$ everywhere & $0.1025$ \\
    \hline
    $80 ~\%$ & $u_0=0$ everywhere & $0.1054$ \\
    \hline
    $90 ~\%$ & $u_0=0$ everywhere & $0.1022$ \\
    \hline
    $50 ~\%$ & $u_0=10$ everywhere & $0.0449$ \\
    \hline
    $50 ~\%$ & $u_0$ as in Equation \ref{eq:sinBCs}  & $0.0442$ \\
    \hline
  \end{tabular}
  \caption{Relative $L_2$ error $\epsilon$ between PANIS and the ground-truth of the non-linear PDE for various out-of-distribution cases.}
  \label{tab:nonLinear}
\end{table}

\subsection{Multiscale PDE}
\label{sub:CGLearningIllustrations}

In this subsection, we demonstrate the multiscale version of the proposed framework (i.e. mPANIS) as described in Section \ref{sec:CGLearning}. 
To this end, we consider the PDE of \refeq{eq:2D_Darcy} and material property fields obtained  for $l=0.05$ (instead of  $l=0.25$ used in the previous examples). 
As a result $c(\bs{s}; \bx)$ exhibits finer scale fluctuations (see indicative examples shown in Figure \ref{fig:xXpairs}). For all the subsequent illustrations presented in subsection \ref{sub:CGLearningIllustrations} we set $\lambda=10^{2.2}$, $N=4096$ and $M=1500$ (see Algorithm \ref{alg:RandomResidualMulti}), while the number of $\bx$-samples used in training was $K=100$.

\begin{figure}[!h]
    \centering
    \includegraphics[width=0.85\linewidth]{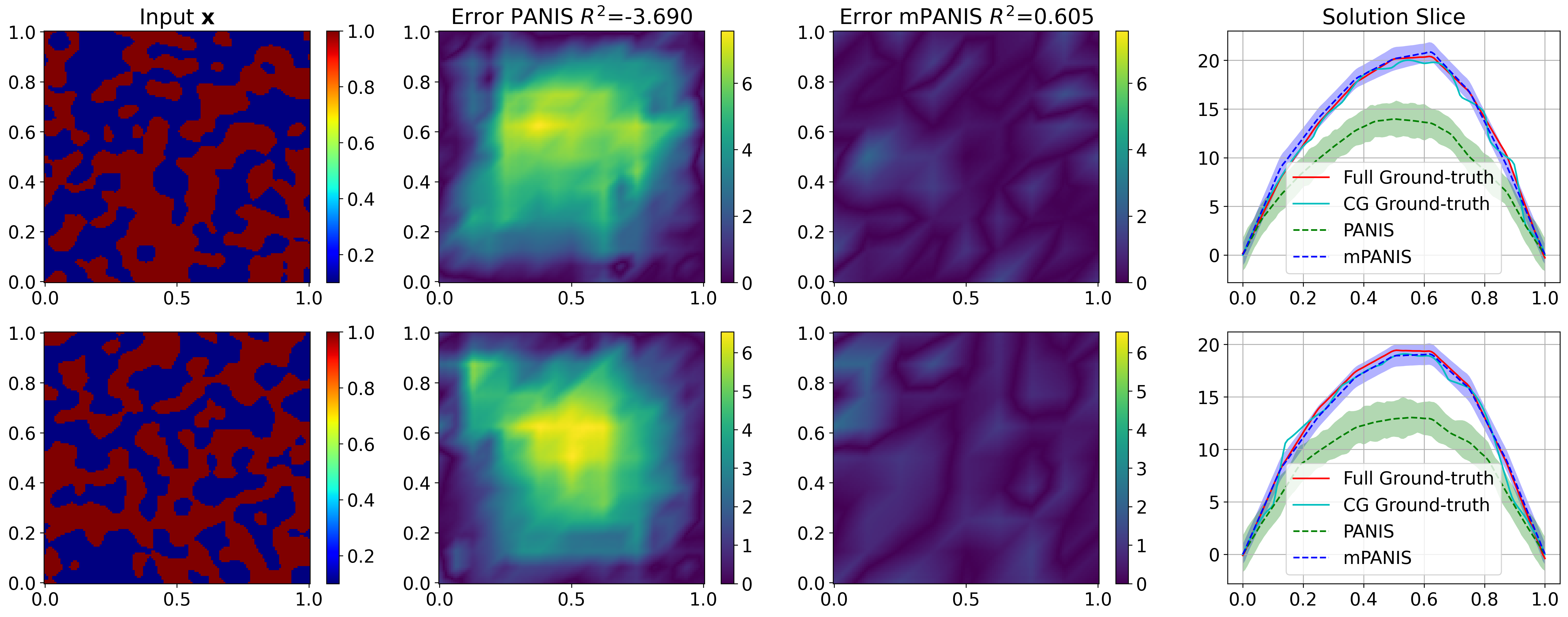}
    \caption{Prediction of PANIS and mPANIS for two indicative inputs ($l=0.05$, $VF=50~\%$ and $u_0=0$).}
    \label{fig:CGLearningWandWO}
\end{figure}

In Figure \ref{fig:CGLearningWandWO}, we show the results obtained by PANIS and mPANIS for indicative microstructures\footnote{Similar results are observed for any such input microstucture} as well as the $R^2$ score over the validation dataset.
One readily notes that despite the fact that the true solution is dominated by low-frequency components, PANIS fails to capture them because it ignores the finer-scale fluctuations. As explained in Section \ref{sec:CGLearning} these might be small but have a significant effect on the corresponding weighted residuals. In contrast, under mPANIS where such fluctuations are accounted for, one is able to capture the low-frequency components of the solution as well probabilistic predictive estimates.

\begin{figure}[!h]
    \centering
    \includegraphics[width=0.85\linewidth]{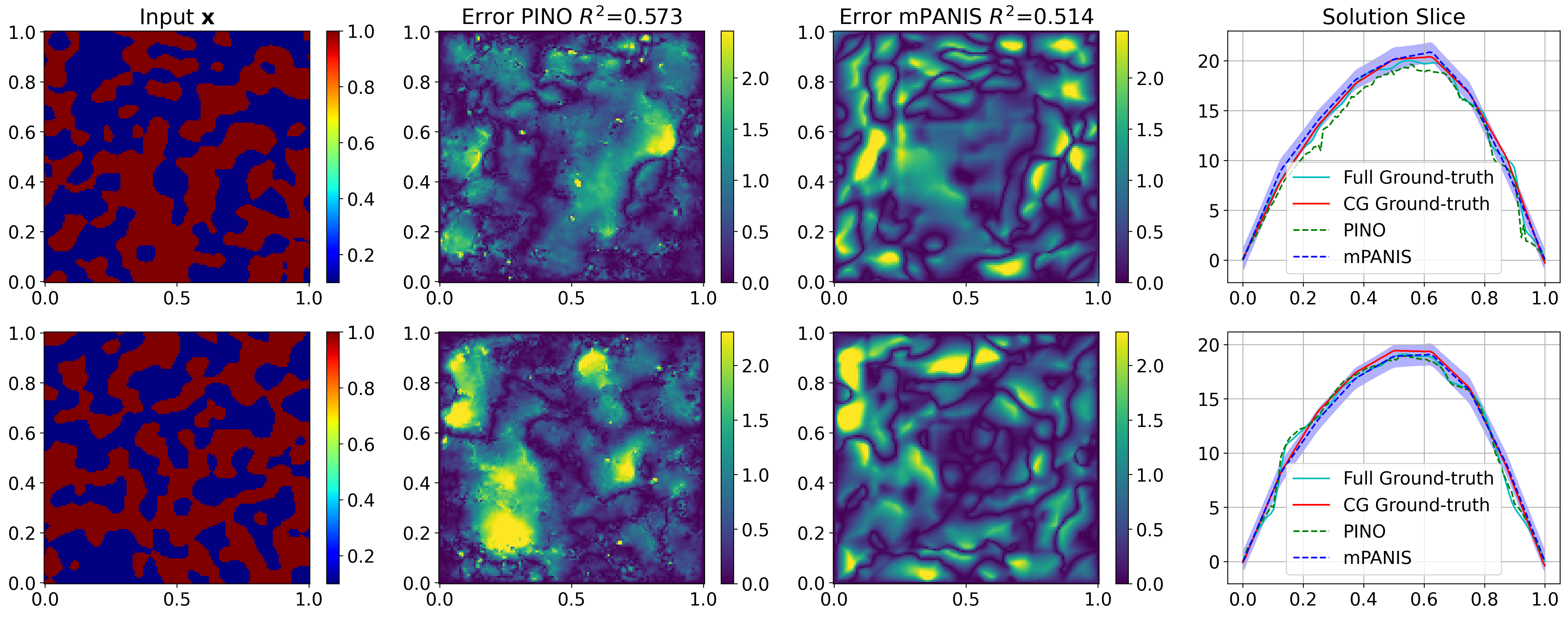}
    \caption{Comparison between PANIS and PINO for conductivity fields with $l=0.05$, $VF=50~\%$ and $u_0=0$ (\textbf{In-distribution predictions}). }
    \label{fig:mPANISVSPINO}
\end{figure}

In Figure \ref{fig:mPANISVSPINO}, we compare the predictive accuracy of mPANIS with PINO. We note that PINO  tries to resolve the full-order solution.  
Despite the fact that mPANIS provides probabilistic solutions and that it  employs roughly {\em two orders of magnitude} fewer parameters and has a training time that is $\approx 3$ times lower (wall clock time in the same machine - see Table \ref{tab:compInput}), the two models have similar accuracy when compared to the full-order solution.

\begin{figure}[!h]
    \centering
    \includegraphics[width=0.5\linewidth]{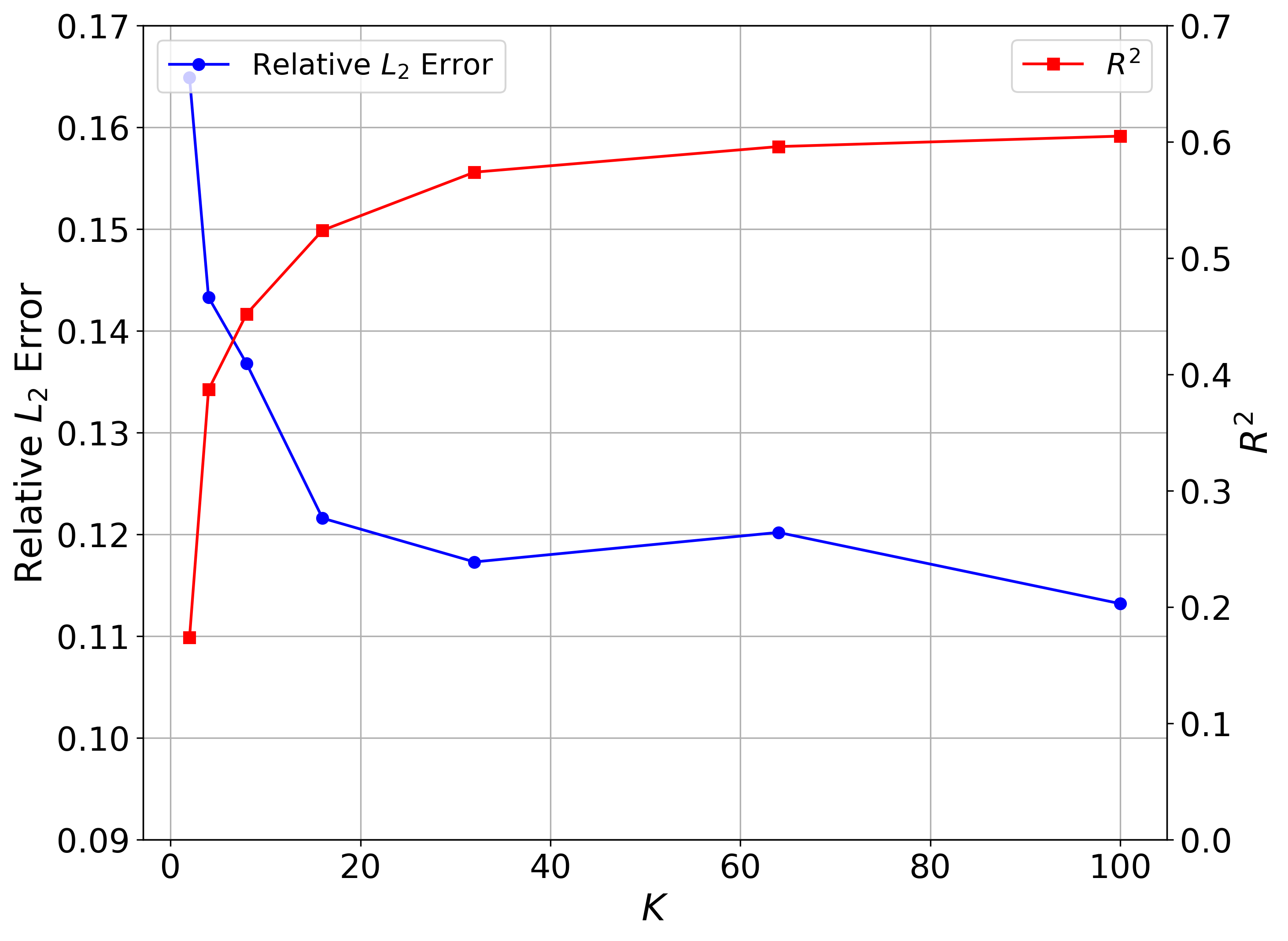}
    \caption{Predictive accuracy of mPANIS in terms of  $R^2$ and relative $L_2$ error $\epsilon$ with respect to the number $K$ of atoms $\bx_k$ (\refeq{eq:pemp}). We observe that mPANIS saturates with fewer than $\sim 50$ such atoms.}
    \label{fig:mPANISSaturation}
\end{figure}

With regards to the number of training parameters and as noted in Section \ref{sec:CGLearning}, this depends on $K$ i.e. the number of $\bx-$samples used in training. We had argued therein that the model architecture is such that it saturates with a small $K$. The validity of this hypothesis is investigated in Figure \ref{fig:mPANISSaturation} which depicts the evolution of the two error metrics employed as a function of $K$. One can see that both stabilize for $K<100$ beyond which little improvement can be expected.

Finally and with regards to out-of-distribution predictions i.e. for different $VF$s or different BCs, mPANIS generally outperforms PINO and in most cases reported in Table \ref{tab:multiscale} by a quite significant margin despite being much more light-weight as explained above.

\begin{table}[!t]
  \centering
  \begin{tabular}{|c|c|c|c|}
    \hline
    Volume Fraction (VF) & Boundary Condition Type & $\epsilon$ (mPANIS) & $\epsilon$ (PINO) \\
    \hline
    $10 ~\%$ & $u_0=0$ everywhere & $0.2240$ & $0.5710$ \\
    \hline
    $20 ~\%$ & $u_0=0$ everywhere & $0.1793$ & $0.4593$ \\
    \hline
    $30 ~\%$ & $u_0=0$ everywhere & $0.1339$ & $0.3321$ \\
    \hline
    $40 ~\%$ & $u_0=0$ everywhere & $0.1131$ & $0.2049$ \\
    \hline
    $50 ~\%$ (training) & $u_0=0$ everywhere & $0.1134$ & $0.1029$ \\
    \hline
    $60 ~\%$ & $u_0=0$ everywhere & $0.1112$ & $0.1078$ \\
    \hline
    $70 ~\%$ & $u_0=0$ everywhere & $0.1450$ & $0.1315$ \\
    \hline
    $80 ~\%$ & $u_0=0$ everywhere & $0.2156$ & $0.1920$ \\
    \hline
    $90 ~\%$ & $u_0=0$ everywhere & $0.3438$ & $0.9850$ \\
    \hline
    $50 ~\%$ & $u_0=10$ everywhere & $0.0634$ & $0.5115$ \\
    \hline
    $50 ~\%$ & $u_0$ as in Equation \ref{eq:sinBCs}  & $0.0671$ & $0.5225$ \\
    \hline
  \end{tabular}
  \caption{Comparison of relative $L_2$ error $\epsilon$ between PANIS and PINO with the ground-truth for various out-of-distribution cases and $l=0.05$.}
  \label{tab:multiscale}
\end{table}

\section{Conclusions}

We have proposed a novel probabilistic, physics-aware framework for constructing surrogates of parametric PDEs.  The most notable and novel characteristics of (m)PANIS are:

\begin{itemize}
    \item it is based on a physics-inspired architecture involving a coarse-grained, implicit solver that introduces a valuable informational bottleneck, which significantly limits the number of trainable parameters and leads to robust generalization.
    \item it efficiently learns the   map between the parametric input and the solution of the PDE by using randomized weighted residuals as informational probes. The trained surrogate provides fully probabilistic predictions that reflect epistemic and model uncertainties,
    \item it can successfully learn the dominant low-frequency component of the solution  in challenging, multiscale  PDEs  by relying solely on the information from the weighted residuals.
    \item it yields probabilistic predictions of similar accuracy with PINO and outperforms it in various out-of-distribution cases while being much more lightweight.
\end{itemize}

The framework introduced provides a multitude of possibilities both in terms of methodological aspects as well as potential applications, the most important of which we enumerate below.

The first pertains to the weighting functions employed. As discussed in section  \ref{sub:probInference}, weighted residuals are used as probes that extract information from the governing PDE. Naturally, and depending on the problem's particulars, their informational content can vary drastically but can nevertheless be quantified in the context of the probabilistic framework adopted through the ELBO. One can therefore envision improving drastically the efficiency of the method by using, adaptively,   weighting functions/residuals that maximize the informational gain. Another opportunity for {\em adaptive learning} arises in the context of mPANIS and the empirical approximation of the input density through $K$ atoms $\bx_k$ (section \ref{sec:CGLearning}). As mentioned therein, the number of training parameters that capture the fine-scale fluctuations of the solution i.e. $\{ \by_{f,k}'\}_{k=1}^K$ is proportional to $K$. One could, therefore, envision progressively increasing $K$ by adaptively incorporating new $\bx_k$'s (and associated $\by_{f,k}'$) on the basis of their ELBO contribution, i.e. the information they furnish. On the methodological front, we finally note the possibility of using a tempering scheme by progressively increasing $\lambda$ (section \ref{infusingVirtualObservables}) or alternative forms of the virtual likelihoods  which could accelerate convergence. 

In terms of applications, an important extension involves dynamical problems. While weighted residuals can be used again to incorporate the PDE in the learning objectives, their dependence on time poses challenges as well as offers several possibilities in terms of enforcement using, e.g. a (linear) multistep method or space-time finite elements \cite{dumont_4d_2018}.
Finally, the most important application pertains to inverse design, i.e. the identification of microstructures $\bx$ that achieve target or extremal properties. The inexpensive surrogate developed, particularly in the multiscale setting, can significantly accelerate this search not only because it can yield accurate predictions faster than the PDE-solver \cite{rixner_self-supervised_2022} but also because, through the use of latent variables ($\bxx, \byy$   in our case) one can obtain derivatives of the response. Due to the discrete nature of the microstructure $\bx$ (and irrespective of the cost of the PDE-solver), these derivatives  are not available in the original formulation of the problem.

\bibliographystyle{unsrt}  
\bibliography{bibliography}

\appendix

\section{Shallow CNN employed for $\bxx_{\bpsi_x}(\bx)$.}
\label{appendix:shallowCNN}

The details for the CNN used in the case of PANIS are shown in Table \ref{tab:CNNPANIS} and in Table \ref{tab:CNNmPANIS} for mPANIS. The only kind of activation function used is the Softplus activation function. These are used after each convolution/deconvolution layer and before the batch normalization layers, except from the last deconvolution layer. The weights of the convolution/deconvolution layers are initialized by using a Xavier uniform distribution \cite{glorot2010understanding}.

\begin{table}[H]
\centering

\begin{tabular}{|l|l|l|l|}
\hline
\textbf{Layers}         & \textbf{Feature Maps}  & \textbf{Height $\times$ Width} & \textbf{Number of Parameters} \\ \hline
Input                       & ---           & 129 $\times$ 129                             &         ---                  \\ \hline
Convolution Layer (k3s2p1)       & 8          & 65 $\times$ 65                             &          80                  \\ \hline
Batch Normalization       & ---          & 65 $\times$ 65                             &            16                \\ \hline
Average Pooling Layer (k2s2)   & ---          & 32 $\times$ 32                             &            ---                \\ \hline
Convolution Layer (k3s1p1)       & 24          & 32 $\times$ 32                             &           1752                 \\ \hline
Batch Normalization       & ---          & 32 $\times$ 32                             &            48                \\ \hline
Average Pooling Layer (k2s2)   & ---          & 16 $\times$ 16                             &               ---             \\ \hline
Deconvolution Layer (k4s1p1)     & 8           & 17 $\times$ 17                             &         3080                   \\ \hline
Batch Normalization       & ---          & 17 $\times$ 17                             &            16                \\ \hline
Deconvolution Layer (k3s1p1)     & ---           & 17 $\times$ 17                             &          73                  \\ \hline
In total     & 40           & ---                             & 5065                        \\ \hline
\end{tabular}
\caption{Layers of the CNN used in PANIS.}
\label{tab:CNNPANIS}
\end{table}

\begin{table}[H]
\centering

\begin{tabular}{|l|l|l|l|}
\hline
\textbf{Layers}         & \textbf{Feature Maps}  & \textbf{Height $\times$ Width} & \textbf{Number of Parameters} \\ \hline
Input                       & ---           & 129 $\times$ 129                             &               ---            \\ \hline
Convolution Layer (k3s1p1)       & 8          & 129 $\times$ 129                             &            80                \\ \hline
Batch Normalization       & ---          & 129 $\times$ 129                             &            16                \\ \hline
Average Pooling Layer (k4s4)   & ---          & 32 $\times$ 32                             &             ---               \\ \hline
Convolution Layer (k3s1p1)       & 16          & 32 $\times$ 32                             &         1168                   \\ \hline
Batch Normalization       & ---          & 32 $\times$ 32                             &            32                \\ \hline
Average Pooling Layer (k2s2)   & ---          & 16 $\times$ 16                             &            ---               \\ \hline
Convolution Layer (k3s1p1)       & 32           & 16 $\times$ 16                             &           4640                 \\ \hline
Batch Normalization       & ---          & 16 $\times$ 16                             &            64                \\ \hline
Average Pooling Layer (k2s2)   & ---          & 8 $\times$ 8                             &              ---              \\ \hline
Deconvolution Layer (k3s1p1)     & 16           & 8 $\times$ 8                             &            4624                \\ \hline
Batch Normalization       & ---          & 8 $\times$ 8                             &            32                \\ \hline
Deconvolution Layer (k4s1p1)     & 8           & 9 $\times$ 9                             &            2056                \\ \hline
Batch Normalization       & ---          & 9 $\times$ 9                             &            16                \\ \hline
Deconvolution Layer (k3s1p1)     & ---           & 9 $\times$ 9                             &          73                \\ \hline
In total     & 80           & ---                             &     12801                       \\ \hline
\end{tabular}
\caption{Layers of the CNN used in mPANIS.}
\label{tab:CNNmPANIS}
\end{table} 
\section{Coarse-to-Fine Output Map $\by(\byy)$}
\label{appendix:ytoYTransform}

In the coarse-grained (CG) model, the solution field in space, i.e. $y_{CG}(\bs{s})$, is represented as:
\be
u_{CG}(\bs{s})=\sum_{J=1}^N Y_J 
 H_J(\bs{s}),
\ee
\noindent where $N=dim(\byy)$ and $H_J$ are the corresponding shape/basis functions. 
If e.g. $\bs{s}_J$ are the nodal points and $H_J$ the usual FE shape functions then $u_{CG}(\bs{s}_J)=Y_J$.

In the reference, fine-grained (FG) model, the solution field, i.e. $u(\bs{s})$, is represented as in \refeq{eq:trialSolRepresentation}:
\be
u(\bs{s})=\sum_{j=1}^n y_j \eta_j(\bs{s}),
\ee

\noindent where $n=dim(\by)$ and $\eta_j$ are the corresponding shape/basis functions. 

The mapping between $\byy$ and $\by$ 
is selected so that:
\be
u_{CG}(\bs{s}) \approx u(\bs{s}).
\ee
One way to quantify the difference/error is e.g.:
\be
E=\frac{1}{2} \int_{\Omega} (u_{CG}(\bs{s}) - u(\bs{s}))^2~d\bs{s}.
\ee
One can readily find that the minimizing $E$ is equivalent to: 
\be
\by=\bs{a}^{-1} \bs{B} \bs{Y},
\label{YtoyTransform}
\ee
where:
\bi
\item $\bs{a}$ is an $n\times n$ matrix  
 with entries $a_{jk}=\int_{\Omega}  \eta_j(\bs{s})  \eta_k(\bs{s})~d\bs{s}$
\item $\bs{B}$ is an $n\times N$ matrix with entries $B_{kJ}=\int_{\Omega}  \eta _k(\bs{s}) ~H_J(\bs{s})  ~d\bs{s}$.\\
\ei

The matrix $\bs{A} = \bs{a}^{-1} \bs{B}$ in \refeq{eq:formOfmeanPrediction} fully defines the coarse-to-fine map and can be pre-computed. 
As a result, there are no parameters to  fine-tune which forces $\bxx$ (on which $\byy$ depends) to attain the physical meaning discussed in section \ref{sec:ApproximatingDensity}. 
\section{ELBO Terms}
\label{appendix:elboTerms}

The computation of the gradient of $\mathcal{F}({\bpsi})$ with respect to $\bpsi$ is based on the auto-ifferentiation capabilities of PyTorch \cite{paszke2019pytorch,bartholomew2000automatic}. 
After each term of the ELBO  has been computed, PyTorch automatically back-propagates through the respective graph to obtain the gradient with respect to the training parameters. 

Regarding the ELBO used for training PANIS, by substituting Equation \eqref{eq:AppoxPostForm} in Equation \eqref{eq:elboRandomResidual}, we obtain the following simplified version of the latter:
\begin{equation}
\begin{array}{ll}
     \mathcal{F} (\bpsi)  \approx  
     -\lambda \frac{N}{M} \sum_{m=1}^M \left< |r_{w_{j_m}}(\by,\bx)| \right> _{q_{\bpsi}(\bx,\by)} & \\
     + \left< \log p(\by | \bx) \right>_{q_{\bpsi}(\bx,\by)}
     - \left< \log q_{\bpsi}(\by|\bx) \right>_{q_{\bpsi}(\bx,\by)}
     ~\text{,where } j_m \sim Cat\left(N,\frac{1}{N}\right).
\end{array}
\label{eq:c1}
\end{equation}

\noindent Given that we have access to $R$ samples $\{\bx_r\}_{r=1}^{R}$ and $\{\by_r\}_{r=1}^{R}$ from the approximate posterior described in Algorithm \ref{alg:RandomResidual}, we approximate the first two terms by Monte Carlo. The second one involves the uninformative multivariate Gaussian prior $\mathcal{N} \left(y| \bs{0},\sigma^2 \bs{I} \right)$, where $\sigma^2 = 10^{16}$ (see subsections \ref{sub:probInference} and \ref{sec:algorithmicImplem}). The last term is the entropy of $q_{\bpsi}(\by|\bx)$  which is calculated  closed form exists since  $q_{\bpsi}(\by| \bx)$ is a multivariate Gaussian distribution. Thus, Equation \ref{eq:c1} can be written (up to a constant) as:

\begin{equation}
\begin{array}{ll}
     \mathcal{F} (\bpsi)  \approx  
     -\lambda \frac{N}{M R} \sum_{m=1}^M \sum_{r=1}^{R} |r_{w_{j_m}}(\by_r,\bx_r)|  & \\
     - \frac{1}{2 R \sigma^2} \sum_{r=1}^{R} \by_r^T \by_r 
     + \frac{1}{2} \log \det \left(\bs{\Sigma}_{\bpsi}\right)
     ~\text{,where } j_m \sim Cat\left(N,\frac{1}{N}\right).
\end{array}
\label{eq:c2}
\end{equation}

In the case of mPANIS described in subsection \ref{sec:algorithmicImplem}, the same procedure was followed as above starting from Equations \ref{eq:elboRandomResidualMultiscale} and \ref{eq:residualTermMulti} by taking into account that:

\begin{itemize}
    \item The priors $p(\by_c | \bx)$ and $p(\by_f^\prime| \bx)$ have the same form as the prior $p(\by | \bx)$ in the case of PANIS previously mentioned.
    \item The entropy term resulting from the degenerate density $\delta\left(\by_f^\prime - \by_{f,k}^\prime \right)$ (see Equation \eqref{eq:qyf}) will have a zero contribution on the ELBO.
\end{itemize}

\noindent By subsampling a subset $\{\bx_r\}_{r=1}^R \subset \{\bx_k\}_{k=1}^K$ and the respective subsets $\{\by_{c,r}\}_{r=1}^R$, $\{\by_{f,r}^\prime\}_{r=1}^R$ the equation with which estimations of the ELBO (up to a constant) are made in the case of mPANIS is:

\begin{equation}
\begin{array}{ll}
     \mathcal{F} (\bpsi)  \approx  
     -\lambda \frac{N}{MR} \sum_{m=1}^M \sum_{i=1}^{R} |r_{w_{j_m}}(\by_{c,r}, \by_{f,r}^\prime, \bx_r)|  & \\
     - \frac{1}{2R\sigma^2} \sum_{i=1}^{R} \left(\by_{c,r}^T \by_{c,r} + \by_{c,r}^{\prime T} \by_{c,r}^\prime \right)
     + \frac{1}{2} \log \det \left(\bs{\Sigma}_{\bpsi}\right)
     ~\text{,where } j_m \sim Cat\left(N,\frac{1}{N}\right).
\end{array}
\label{eq:c3}
\end{equation}

\noindent As one can observe from Equations \ref{eq:c2} and \ref{eq:c3}, when $\sigma$ is big (as in the cases employed), the second term of the ELBO is approximately equal to zero, and thus it could be completely ignored.

\end{document}